\documentclass[journal]{IEEEtran}
\usepackage{cite}
\usepackage{amsmath,amssymb,amsfonts}
\usepackage{graphicx}
\usepackage{textcomp}
\usepackage{graphicx}
\usepackage{amsmath}
\usepackage{amssymb}
\usepackage{booktabs}
\usepackage{multirow}
\usepackage{xcolor}
\usepackage{indentfirst}
\usepackage{newfloat}
\usepackage{listings}
\usepackage{amsmath}
\usepackage{amssymb}
\usepackage{algorithm}
\usepackage{bbm}

\usepackage{setspace}

\usepackage{wrapfig}
\usepackage{url}
\usepackage{soul}

\newcommand{\blue}[1]{{#1}}
\newcommand{\BLUE}[1]{{#1}}
\newcommand{\stdev}[1]{\tiny $\pm$#1}

\usepackage[caption=false,font=footnotesize,labelfont=rm,textfont=rm]{subfig}
\usepackage{threeparttable}
\usepackage{multirow}
\usepackage{enumitem}
\usepackage{makecell}
\usepackage{amssymb}
\usepackage{hyperref}
\begin{document}

\title{Balancing Multi-Target Semi-Supervised Medical Image Segmentation with Collaborative Generalist and Specialists}
\author{You Wang, Zekun Li, Lei Qi, Qian Yu, Yinghuan Shi, Yang Gao
%\thanks{This work was supported by the NSFC Program (62222604, 62206052, 62192783), China Postdoctoral Science Foundation (2021M690609), Jiangsu Natural Science Foundation (BK20210224), CCF-Lenovo Blue Ocean Research Fund, High Level Scientific Research Project Cultivation Fund (2019GSPGJ07), and Discipline Talent Team Cultivation Program of Shandong Women's University (1904).}
\thanks{You Wang, Zekun Li, Yang Gao, and Yinghuan Shi are with the State Key Laboratory for Novel Software Technology, Nanjing University, China. They are also with National Institute of Healthcare Data Science, Nanjing University, China. (E-mail: wangu@smail.nju.edu.cn, lizekun@smail.nju.edu.cn, gaoy@nju.edu.cn, syh@nju.edu.cn)}
\thanks{Lei Qi is with the School of Computer Science and Engineering, and the Key Lab of Computer Network and Information Integration (Ministry of Education), Southeast University, China. (E-mail: qilei@seu.edu.cn)}
\thanks{Qian Yu is with the School of Data and Computer Science, Shandong Women’s University, China. (E-mail: yuqian@sdwu.edu.cn)}
\thanks{The corresponding author of this work is Yinghuan Shi.}
}
\maketitle

\begin{abstract}
Despite the promising performance achieved by current semi-supervised models in segmenting individual medical targets, many of these models suffer a notable decrease in performance when tasked with the simultaneous segmentation of multiple targets.
A vital factor could be attributed to the imbalanced scales among different targets: during simultaneously segmenting multiple targets, large targets dominate the loss, leading to small targets being misclassified as larger ones. To this end, we propose a novel method, which consists of a Collaborative Generalist and several Specialists, termed CGS. It is centered around the idea of employing a specialist for each target class, thus avoiding the dominance of larger targets.
The generalist performs conventional multi-target segmentation, while each specialist is dedicated to distinguishing a specific target class from the \blue{remaining} target classes and the background.
Based on a theoretical insight, we demonstrate that CGS can achieve a more balanced training.
Moreover, we develop cross-consistency losses to foster collaborative learning between the generalist and the specialists. Lastly, regarding their intrinsic relation that the target class of any specialized head should belong to the \blue{remaining} classes of the other heads, we introduce an inter-head error detection module to further enhance the quality of pseudo-labels. Experimental results on three popular benchmarks showcase its superior performance compared to state-of-the-art methods. Our code is available at \href{https://github.com/wangyou0804/CGS}{\colorbox{pink}{{\texttt{https://github.com/wangyou0804/CGS}}}}.

\end{abstract}


\section{Introduction}
\IEEEPARstart{A}{iming}  to provide pixel-wise predictions to medical images, \emph{e.g.}, computed tomography (CT) 
and magnetic resonance imaging (MRI), medical image segmentation is a longstanding yet significant task in computer vision and pattern recognition \cite{ramesh2021review, chen2018drinet}. However, the conventional fully-supervised medical image segmentation method heavily relies on meticulously annotated images by experienced radiologists, which considerably hinders its development in clinical scenarios, since annotating a sufficient number of medical images is expensive, subjective, and even infeasible. To tackle this issue, semi-supervised medical image segmentation (SSMIS) is widely employed \cite{nie2018asdnet, li2020self}, which aims to utilize a small amount of labeled data along with a large amount of unlabeled data to automatically provide pixel-wise predictions for medical images, which alleviates the scarcity of label.

Existing SSMIS methods generally rely on consistency regularization \cite{luo2021semi, bortsova2019semi} and pseudo-labeling \cite{chaitanya2023local, zhang2022boostmis, li2022pln, ma2025steady}. Despite their success, we have noticed that most of these methods neglect the interference between different categories\footnote{We exchangeably use \emph{target}, \emph{class} and \emph{category} in this paper.}. \blue{When employing them to simultaneously segment different targets (\emph{i.e.}, organs, tissues, \emph{etc.}), these previous methods would suffer from performance degradation to some extent. This occurs due to mutual interference among categories caused by class imbalance, which can be categorized into occurrence imbalance and scale imbalance.} Thus, this research is conducted to investigate the crucial cause of such performance degradation.

\begin{figure}[t]
    \centering
    \includegraphics[width=\linewidth]{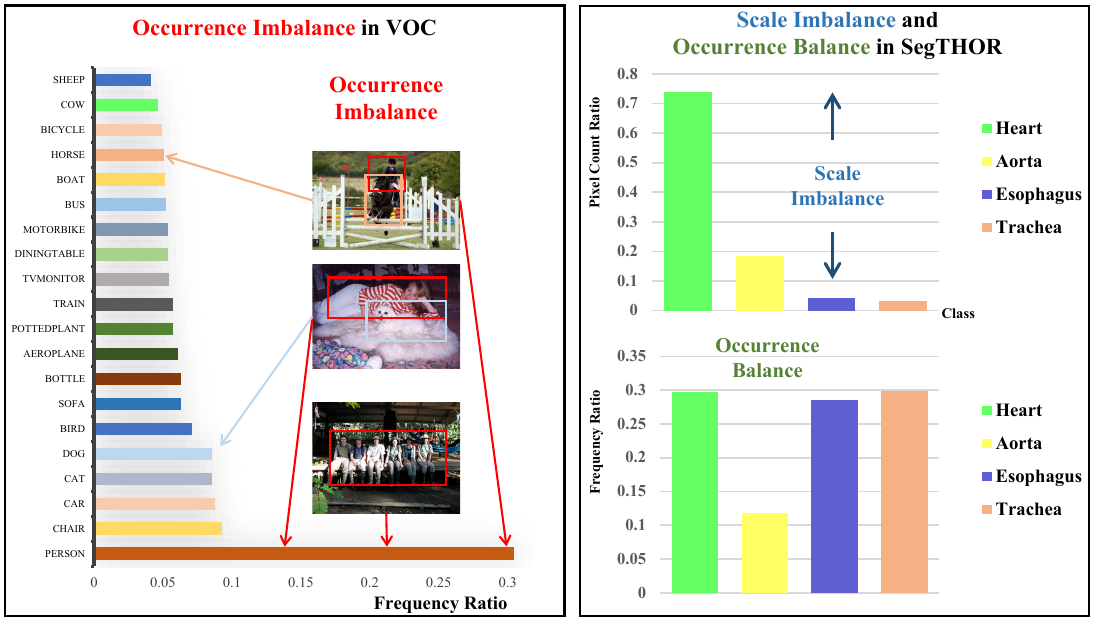}
    \vspace{-20pt}
    \caption{\blue{The left figure illustrates the occurrence imbalance in the VOC dataset, where the frequency of class occurrences is uneven. The right figure shows the scale imbalance in the SegTHOR dataset, where the occurrence between different class is relatively balanced.}}
    \label{fig:intro}
    \vspace{-20pt}
\end{figure}

Considering that a segmentation task can be viewed as a pixel-level classification task, it is intuitive that imbalanced numbers of pixels in different classes can diminish segmentation performance, which refers to class imbalance.
The imbalanced segmentation problem in natural images has been explored in prior studies~\cite{ zhou2023dynamic}, where the class imbalance issue primarily arises from the uneven occurrence of different targets. However, in medical image segmentation, multiple organs or tissues often anatomically appear simultaneously, thus the issue stems from scale imbalance rather than occurrence imbalance. As a consequence, existing imbalanced segmentation methods are less applicable to medical tasks. \blue{For example, Precise delineation of organs at risk (OAR) is a critical task in radiotherapy treatment planning (RTP), as it aims to deliver a high dose to the tumor while minimizing exposure to healthy tissues~\cite{fechter2017esophagus}. 
When segmenting multiple OAR organs simultaneously, the scale imbalance between different organs can lead to poorer segmentation performance for small organ (\emph{e.g.}, esophagus). }

To this end, we focus on the \textbf{scale imbalance} issue in multi-target SSMIS~\cite{gibson2018automatic, lei2020deep}. \BLUE{Note that, scale imbalance is not a new concept to replace the very common notion of class imbalance~\cite{li2019overfitting}, while it arises from  the intrinsic characteristics of medical images and has not yet received widespread attention.}
\BLUE{Scale imbalance is a specific form of class imbalance that emphasizes the unequal distribution of targets of varying sizes across the entire dataset.
It differs from another form of class imbalance, known as occurrence imbalance, which deals with the unequal frequency of class appearances.}
% \BLUE{Most researches about segmentation addressing the imbalance issue likely refer to the overall imbalance problem as discussed in this paper. We argue that occurrence imbalance and scale balance should be studied separately and we focus on the more fine-grained scale imbalance as the starting point for our research.}
\blue{To be more specific, in Fig.~\ref{fig:intro}, 
the left figure depicts the occurrence imbalance in the natural image dataset, VOC~\cite{everingham2010pascal}, characterized by uneven class frequencies. In contrast, the right figure illustrates the scale imbalance in the medical image dataset, SegTHOR~\cite{lambert2020segthor}, where class occurrences are relatively balanced but the sizes of the objects vary significantly across classes.}

Basically, we notice that such an issue is a previously underestimated factor in multi-target segmentation caused by the substantial variations in scale among different targets in the human body. Under scale imbalance conditions, large-scale targets prevailing in the foreground might greatly result in the model tending to prioritize the segmentation of large-scale targets, leading to the misidentification of smaller targets and backgrounds as larger ones. \blue{More specially, in the SegTHOR dataset, this scale imbalance causes the model to focus more on the larger target (heart) during segmentation, resulting in misclassification of background regions as the heart.} Note that, this scale imbalance issue could also occur in fully supervised segmentation, whereas an ample supply of labeled data can mitigate the impact of scale imbalance, which could not be realized in the semi-supervised scenario.

A feasible solution to tackle scale imbalance challenge involves dividing the multi-target segmentation task into subtasks dedicated to each specific target class. This solution, akin to the one-vs-rest approach, avoids the interference among different categories, which is attempted by employing expert decoders for individual classes in semantic segmentation \cite{liu2023self}. However, such a straightforward solution may lead to excessive independence among the individual subtasks, resulting in the absence of correlation information among different targets. Thus, it is natural to consider \emph{whether we can jointly optimize the general segmentation for multiple different targets and the specialized segmentation for each individual target.} To achieve this goal, we propose to \textbf{C}ollaboratively train with both a \textbf{G}eneralist and \textbf{S}pecialists, termed as CGS.

\begin{figure}[t]
    %\vspace{-20pt}
    \centering
    \includegraphics[width=0.8\linewidth]{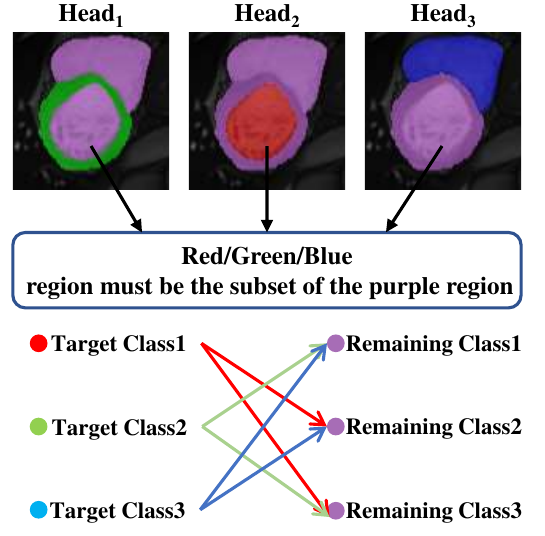}
    \vspace{-10pt}
    \caption{\blue{Green, red, and blue respectively represent the target classes of three segmentation heads, while purple represents the \blue{remaining} classes. Taking red as an example, the red region in the middle image should belong to the purple regions in the two side images. The \blue{remaining} class consists of several target classes. Best version in color.}}
    \label{fig:intro2}
    \vspace{-20pt}
\end{figure}

In pursuit of this goal, we introduce a collaborative training strategy comprising a generalist and several specialists. Specifically, the generalist is equipped with a segmentation head to perform the segmentation of multiple targets. 
Meanwhile, the specialists consist of multiple class-specific segmentation heads. 
To prevent specialists from overlooking the existence of other categories,
each head is dedicated to segmenting the corresponding \emph{target class}, \emph{\blue{remaining} classes}, and \emph{background}. In this strategy, each class contributes to the training of other specialized segmentation heads as the \blue{remaining} classes. \blue{To be more specific, in Fig.~\ref{fig:intro2}, during the training of the segmentation head for class 1 (red), the rest/remaining classes (purple) are composed of class 2 (green) and class 3 (blue).} This ensures that all categories contribute to training with a relatively balanced representation at the pixel level, effectively alleviating the issue of scale imbalance. Thanks to our divide-and-conquer strategy, a more balanced training process is achieved. A theoretical insight about the more balanced training proportion of different targets is included in this paper.

\blue{Concurrently, in order to achieve consensus between the generalist and the specialists, we adopt a bidirectional consistency loss and a consensus-based consistency loss.} Through the collaborative training of both participants, their segmentation capabilities are mutually reinforced and tend to converge. To ensure parameter efficiency, the specialists are discarded in inference, maintaining the same parameter size as conventional methods, \emph{e.g.}, UNet~\cite{ronneberger2015u}.

Moreover, in order to further exploit the relationships among all the segmentation heads of the specialists, we introduce an Inter-Head Error Detection (IHED) module. We observe that, for any two specialized segmentation heads, a pixel in the target class should belong to the \blue{remaining} classes of the other head, and vice versa.
For instance, in the ACDC dataset, when the head is dedicated to segmenting the right ventricle as the target class, the left ventricle is categorized within the \blue{remaining} classes (see Fig.~\ref{fig:intro2}). Consequently, it is reasonable to leverage this motivation to detect pixel-level errors in the pseudo-labels generated from the unlabeled data. 

We summarize the contributions of this paper as follows:
\begin{itemize}
    \item To the best of our knowledge, we are the first to investigate the challenge of scale imbalance in multi-target SSMIS. 
    \item We propose a novel collaborative training framework that involves a generalist and several specialists to achieve a more balanced training, without adding additional parameters during inference.
    \item Building upon the specialists, we propose the Inter-Head Error Detection (IHED) module to enhance the quality of pseudo-labels from the unlabeled data. 
    \item Additionally, we incorporate bidirectional the consistency loss and consensus-based consistency loss to achieve consensus between the generalist and the specialists.
\end{itemize}

\blue{We extensively evaluate our method on ACDC, SegTHOR and Synapse, surpassing the state-of-the-art (SOTA) methods, \emph{e.g.}, on ACDC, DSC of 87.27\% \emph{vs.} 88.83\% with 3 labeled scans and DSC of 88.89\% \emph{vs.} 89.83\% with 7 labeled scans.}

\section{Related Work}

\begin{figure*}[t]
    \centering
    \includegraphics[width=0.95\linewidth]{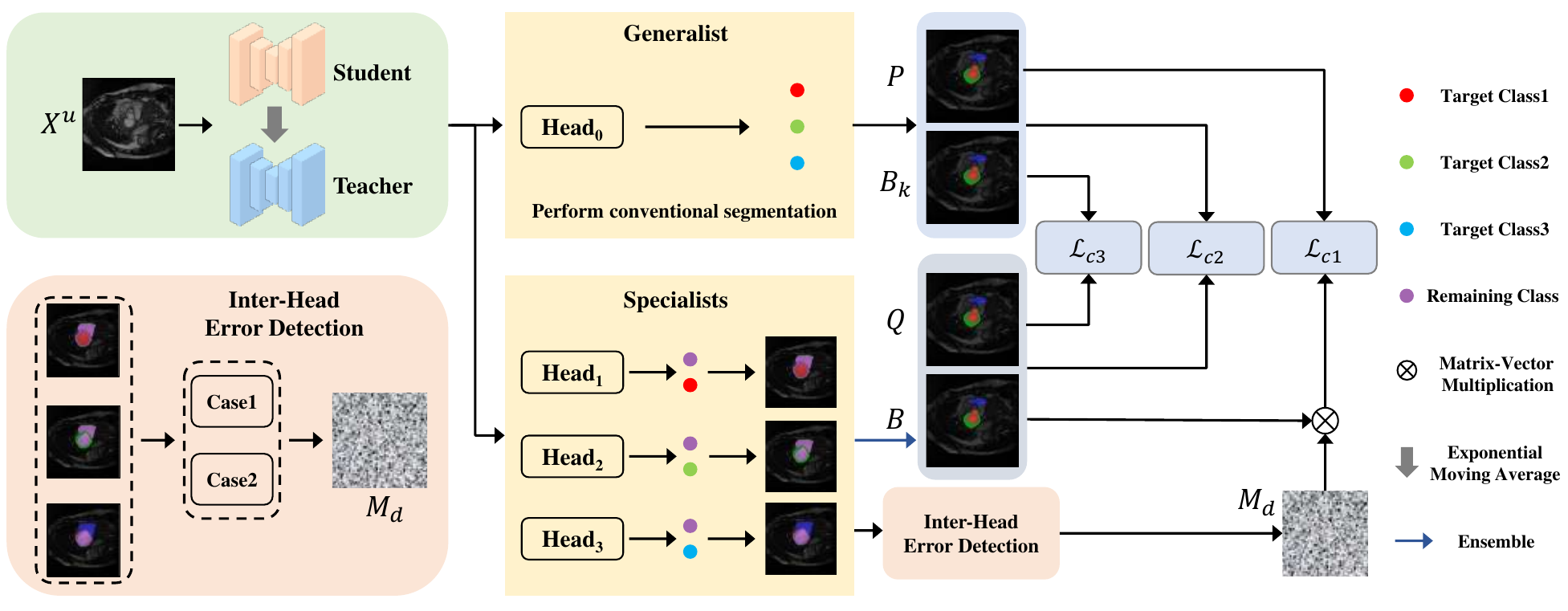}
    % \vspace{-10pt}
    \caption{\blue{The pipeline of our proposed CGS (illustrated using ACDC as an example). The framework comprises the generalist as the general branch and the specialists as the multi-head specialized branch. Building upon this, the cross-branch consistency losses are computed and the inter-head error detection module is achieved.
    $P$ represents predictions from the general branch for unlabeled data $X^u$, while $Q$ denotes the predictions from the specialized branch. $B_k$ is the pseudo-label from the general branch for the specialized branch, while $B$ is the pseudo-label from the specialized branch for the general branch. Additionally, $M_d$ is employed to represent the error detection matrix. }}
    \label{fig:pipeline}
    \vspace{-10pt}
\end{figure*}

\textbf{Semi-supervised medical image segmentation.} Due to the expertise required and the time-consuming process of annotating medical images, there have been significant efforts towards semi-supervised medical image segmentation (SSMIS) in recent years. Contrastive learning-based methods \cite{lou2023min, pandey2021contrastive} learn representations that maximize the similarity among negative pairs and minimize the similarity among positive pairs. Simultaneously, numerous studies have expanded upon the Mean Teacher \cite{tarvainen2017mean} framework with consistency regularization. Yu \emph{et al.} estimates the uncertainty of pseudo-labels with Monte Carlo Dropout \cite{kendall2017uncertainties} in UA-MT \cite{yu2019uncertainty}. CoraNet \cite{shi2021inconsistency} estimates uncertainty by capturing the inconsistent prediction between multiple cost-sensitive settings. Luo \emph{et al.} proposes a novel method of estimating uncertainty by capturing the inconsistent prediction between multiple cost-sensitive settings in DTC \cite{luo2021semi}. These methods generate high-quality pseudo-labels from the perspective of uncertainty estimation. Additionally, some methods impose constraints on pseudo-labels from a geometric perspective, \emph{i.e.}, SASSNet \cite{li2020shape}. Moreover, Bai \emph{et al.} employs bidirectional CutMix on 2D slices in BCP \cite{bai2023bidirectional}, while Chen \emph{et al.} leverages bidirectional cube partition techniques on 3D volumes, resulting in significant advancements in data augmentation methodologies in MagicNet \cite{chen2023magicnet}. \BLUE{Additionally, FixMatchSeg \cite{upretee2022fixmatchseg} is a direct extension of FixMatch~\cite{sohn2020fixmatch} to medical image segmentation, which focuses on enforcing consistency loss between weakly and strongly augmented unlabeled data.}
% Furthermore, recent studies \cite{cai2023orthogonal, li2022pln} employ sparse annotations on 3D volumes, utilizing fewer labeled data as the SSMIS variant. 
\blue{However, these methods often overlook the issue that, in situations
where labeled data is extremely limited, scale imbalance can cause small-scale targets to be underrepresented during
training. This may weaken the ability of model to effectively segment small-scale targets. This can result in poor segmentation quality of small target regions in the pseudo-labels generated from unlabeled data, which in turn leads to error accumulation and a degradation in model performance.} \blue{Many methods have been developed for semi-supervised natural image segmentation~\cite{olsson2021classmix}.}

% \subsection{}
\textbf{Multi-target SSMIS.} Due to the simultaneous presence of numerous targets or tissues in CT or MRI scans, multi-target segmentation is a crucial task \cite{gibson2018automatic, lei2020deep}. 
\blue{Existing multi-target medical image segmentation \cite{conze2021abdominal} methods encounter bottlenecks when sufficient labeled data is not available and class imbalance affects the segmentation performance of model.}
\blue{The class imbalance issue in Multi-target SSMIS encompasses both class occurrence imbalance and scale imbalance, both of which impact segmentation performance. Scale imbalance refers to the disparities in the spatial distribution of different classes within the dataset, while occurrence imbalance focuses on the unequal frequency of class appearances. }
However, very few semi-supervised methods have concentrated on multi-target segmentation. To be more specific, DMPCT \cite{zhou2019semi} employs a 2D co-training framework to aggregate multi-planar features, while UMCT \cite{xia2020uncertainty} emphasizes multi-view consistency on unlabeled data. MagicNet \cite{chen2023magicnet} encourages unlabeled data to acquire organ semantics from the labeled data in relative locations. Although these methods have demonstrated promising performance in multi-target SSMIS, they overlook the issue of scale imbalance in semi-supervised medical image segmentation. 
\blue{
In semi-supervised semantic segmentation, USRN~\cite{guan2022unbiased} addresses class imbalance through clustering techniques. However, this approach is not well-suited for medical image segmentation, where the number of categories is typically limited. Compared to natural images, medical images present unique challenges such as blurred boundaries, noise interference, and intricate anatomical structures. Consequently, many segmentation methods developed for natural images struggle to deliver strong performance on medical image tasks.}
Moreover, the performance of SAM~\cite{kirillov2023segment} and SAM-based methods~\cite{wu2023medical,mazurowski2023segment,li2025stitching} exhibits significant variability depending on the dataset and specific task. 
Particularly in multi-target medical image segmentation, the segmentation results are notably suboptimal, as the majority of their training data is derived from natural images, whereas medical images typically exhibit different statistical characteristics and textures. \BLUE{For example, we use bounding boxes as prompts to input into the MedSAM \cite{ma2024segment} model for predictions on the ACDC and SegTHOR test sets. The predicted segmentation results yield dice coefficients of only 50.59\% and 47.56\%, respectively. Furthermore, training SAM \cite{kirillov2023segment} and MedSAM requires substantial data, memory, and time. As a result, large models still exhibit certain limitations when applied to specific medical image segmentation tasks.}

\textbf{Remark.} When using the current SSMIS method to segment multiple targets simultaneously, previous approaches tend to experience reduced performance because different categories can mutually interfere with each other, primarily due to scale imbalance. \blue{
Thus, there is a need to explore a novel approach tailored for multi-target semi-supervised medical image segmentation, aiming to enhance the segmentation efficacy of model.}

\section{Method}
\subsection{Notations and Overview}
\blue{We define each 2D slice of medical images as $X \in \mathbbm{R}^{H \times W}$, where $H$ and $W$ denote the height and width of an image, respectively. The goal of multi-target semi-supervised medical image segmentation is to find the semantic label $\tilde{Y} \in \{0,1,..., K\}^{H \times W}$ for each slice $X$, where $K$ denotes the number of classes,  $0$ denotes background and other numbers denote specific classes.} The training set $\mathcal{D}$ consists of two subsets: $\mathcal{D} = \mathcal{D}^l \cup \mathcal{D}^u$, where $\mathcal{D}^l = \{(X_i^l, Y_i^l)\}_{i=1}^M$ and $\mathcal{D}^u = \{X_i^u\}_{i=1}^N$ ($M \leq N)$. The training images $X^l$  in dataset $\mathcal{D}^l$ are provided with pixel-wise annotations $ Y^l $, while $ X^u $ in dataset $ \mathcal{D}^u $ are not.
 In the context of $X$, $Y$ and $P$, the superscript $l$ and $u$ denote labeled and unlabeled, respectively.

In this study, we utilize the teacher-student framework with weak-to-strong consistency as the foundation. The overall framework of the proposed CGS is shown in Fig.~\ref{fig:pipeline}.  
Technically, CGS adopts a dual-branch framework, with the generalist implemented as the general branch and the specialists as the multi-head specialized branch.

\blue{Overall, we first present the basic teacher-student framework for semi-supervised medical image segmentation, which serves as the general branch. Building on this foundation, we introduce a multi-head specialized branch comprising multiple specialized heads designed to segment the target class, remaining classes, and background. To enhance collaboration between the two branches, we establish a cross-branch consistency loss that facilitates mutual learning. Furthermore, we propose the Inter-Head Error Detection (IHED) module, which improves pseudo-label quality by analyzing conflicts among the specialists. Finally, we address the memory overhead introduced by our framework during the inference phase.} 
% \subsection{General Branch}
\subsection{Generalist as General Branch}
In our method, we adopt a teacher network $\mathcal{\hat{F}}$ and a student network $\mathcal{F}$. The student network is optimized by stochastic gradient descent, and the teacher network is updated with the exponential moving average of the student network.  

The teacher network generates pseudo-labels for unlabeled data, which are then fed into the student network. To improve the generalization capability of model, we apply various data augmentation techniques to both labeled and unlabeled data. Concretely, we adopt cropping, flipping, and rotation as weak augmentation $\mathcal{A}_w(\cdot)$ to the input $X$ of the teacher network while color enhancement (brightness, contrast, and saturation) and CutMix \cite{you2022simcvd} as strong augmentation $\mathcal{A}_s(\cdot)$ to the input $X^u$ of the student network. All the labeled images $X^l$ are fed into the student network for training in a supervised manner with $\mathcal{A}_w(\cdot)$, while unlabeled data has been augmented with strong augmentation $\mathcal{A}_s(\cdot)$. In this way, we can formulate the probability maps from student network $\mathcal{F}$  as:
% Their probability maps are computed as:
\begin{equation}
%\vspace{-2pt}
\begin{aligned}
    P_w^l&=\sigma\big(\textbf{Seg}\big(\mathcal{F}(\mathcal{A}_w(X^l))\big)\big),\\ P_s^u&=\sigma\big(\textbf{Seg}\big(\mathcal{F}(\mathcal{A}_s\left(X^u\right))\big)\big),
\end{aligned}
\end{equation}
% \begin{equation}
%     P_s^u=\sigma(\textbf{Seg}(\mathcal{F}(\mathcal{A}_s(X^u)))),
% \end{equation}
where $\sigma(\cdot)$ denotes the softmax layer, $\textbf{Seg}(\cdot)$ represents the general segmentation head comprising a convolutional layer, and the predicted probability map of unlabeled images from teacher network $\mathcal{\hat{F}}$ can be computed as follows:
\begin{equation}
\label{eq3}
    \hat{P}_w^u = \sigma \big(\textbf{Seg}\big(\mathcal{\hat{F}}\left(\mathcal{A}_w\left(X^u\right)\right)\big)\big).
\end{equation}

Consequently, we derive the pseudo-labels for unlabeled images from the teacher network. This process enables the definition of the supervised loss $\mathcal{L}_{sup}$ and the unsupervised loss $\mathcal{L}_u$ as follows:
\begin{equation}
    \mathcal{L}_{sup} = \mathcal{H}\left(P_w^l, Y^l\right),
\end{equation}
\begin{equation}
    \mathcal{L}_u = \mathbbm{1}\big(\textbf{max}(\hat{P}^u_w) > \tau\big) \mathcal{H}\big(P_s^u, \textbf{argmax}(\hat{P}^u_w)\big),
\end{equation}
% \begin{equation}
%     \mathcal{L}_u = \mathbbm{1}(\textbf{max}(\hat{P}^u_w) > \tau) \mathcal{H}(P_s^u, \textbf{argmax}(\hat{P}^u_w)),
% \end{equation}
where $\mathbbm{1}(\cdot)$ represents an indicator function. $\tau$ is a predefined confidence threshold to filter uncertain pseudo-labels, and $\mathcal{H}(\cdot)$ encompasses the losses in semi-supervised medical image segmentation, comprising cross-entropy loss and dice loss:

\begin{equation}
    \mathcal{H}(\hat{Y}, Y) = \mathcal{L}_{ce}(\hat{Y}, Y) + \mathcal{L}_{dice}(\hat{Y}, Y),
\end{equation}
 \blue{where $\hat{Y}$ and $Y$ represent the prediction of model and the ground truth, respectively, with $\mathcal{L}_{ce}$ denoting the cross-entropy loss and $\mathcal{L}_{dice}$ representing the dice loss.}

% \subsection{Multi-head Branch}
\subsection{Specialists as Multi-Head Specialized Branch }
\label{sec::3.2}
In order to mitigate the scale imbalance issue in SSMIS, we are exploring the insight of assigning a specialized segmentation head for each class. 
\blue{Unlike conventional Binary Segmentation Heads (BSH), which solely focus on the binary segmentation task (target class and background), we are exploring the design of a segmentation head that emphasizes the interrelationship between different classes.}
To strengthen the relationship between the target class and other classes within each head, the task of the head is defined as classified into three categories: \textbf{target}, \textbf{\blue{remaining} classes}, and \textbf{background} (see Fig.~\ref{fig:pipeline}).

\begin{figure}[h]
    \centering
    \includegraphics[width=\linewidth]{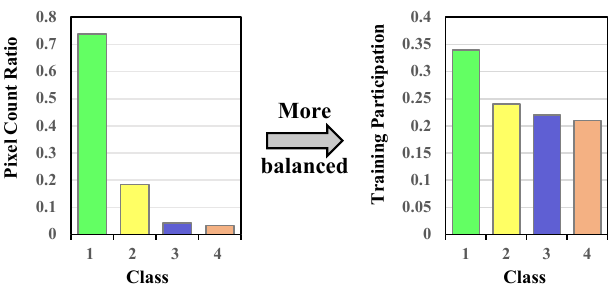}
    \vspace{-10pt}
    \caption{\blue{The left figure shows the training participation proportions of different classes using the conventional training method, while the right figure illustrates the corresponding proportions in our method.}} 
    \label{fig:balance}
    % \vspace{-30pt}
\end{figure}

\textbf{Label Redefinition.}  In this purpose, we rebuild the label $Y \in \{0,1, \dots, $ $K\}^{H \times W}$ into $Z_i \in \{0,1,2\}^{H \times W}$, $i \in \{1,2,...K\}$ as follows:
$$
Z_i^{(x, y)} = 
\begin{cases}
    0 & \text{if } Y^{(x, y)} = 0 \\
    1 & \text{if } Y^{(x, y)} = i \\
    2 & \text{otherwise}
\end{cases}
$$
where $Z_i^{(x, y)}$ and $Y^{(x, y)}$ represents the pixel located in the $x$-th row and the $y$-th column. The values $0$, $1$, and $2$ correspond to \textbf{background}, \textbf{target class}, and \textbf{\blue{remaining} classes}, respectively. \blue{In Fig.~\ref{fig:pipeline}, we provide a visualization label redefinition. More specially, during the training of the segmentation head for class 1 (red), the remaining classes (purple) are composed of class 2 (green) and class 3 (blue). The training procedure for the other classes follows the same paradigm.} \blue{In addition, on the left of Fig.~\ref{fig:balance}, we show the pixel count ratio of each class in the entire dataset, while the right side displays the training participation after applying our pretext task. As demonstrated in the figure, our strategy effectively mitigates the imbalance in training participation that is present in the original dataset.}

\blue{Thanks to the label redefinition strategy, we design $K$ additional pretext tasks for a more balancing training. Since medical image segmentation typically involves a limited number of categories and we avoid adding complex modules, the additional computational cost remains manageable.
 }

Note that, redefining labels does not lead to gradient conflicts, as all segmentation heads can leverage shared features from the encoder and decoder. On the contrary, this collaborative utilization of features ensures a cohesive learning process across different segmentation tasks, further enriching the understanding of complex visual patterns. Moreover, by performing segmentation tasks at varying scales based on the shared features, the model enhances its segmentation accuracy, thereby improving its overall feature extraction capabilities.

Furthermore, to enhance diversity among segmentation heads, as well as between the multi-head specialized branch and the general branch of the network, we introduce a projector before the segmentation head. Each projector consists of a convolutional layer with the same number of input and output channels, followed by a BN layer and a ReLU layer. The probability maps from the $k$-th $(k=1,2,..., K)$ segmentation head can be formulated as: 
% %\vspace{-5pt}
\begin{equation}
    Q_{k,w}^l=\sigma \big(\textbf{Seg}_k\big(\textbf{Proj}_k\big(\mathcal{F}(\mathcal{A}_w(X^l))\big)\big)\big), 
\end{equation}
\begin{equation}
    Q_{k,s}^u=\sigma \big(\textbf{Seg}_k\big(\textbf{Proj}_k\big(\mathcal{F}(\mathcal{A}_s(X^u))\big)\big)\big),
\end{equation}
where $\textbf{Seg}_k(\cdot)$ and $\textbf{Proj}_k(\cdot)$ are the $k$-th segmentation head and the corresponding projector, respectively. Similar to Eq.~(\ref{eq3}), we can get the pseudo-label probability map from the $k$-th specialized segmentation head:
\begin{equation}
    \hat{Q}_{k,w}^{u} = \sigma \big(\textbf{Seg}_k\big(\textbf{Proj}_k\big(\mathcal{\hat{F}}(\mathcal{A}_w(X^u))\big)\big)\big).
\end{equation}

Thus, we can train the specialized segmentation heads with supervised loss $\mathcal{L}_{sup}^\text{MH}$ and unsupervised loss $\mathcal{L}_u^\text{MH}$:
%\vspace{-5pt}
\begin{equation}
\begin{gathered}
    \mathcal{L}_{sup}^\text{MH} = \frac{1}{K}\sum_{k=1}^K\mathcal{H}\left(Q_{k,w}^{l}, Z^{l}_k\right), \\
    \mathcal{L}_u^\text{MH} = \frac{1}{K}\sum_{k=1}^K \mathbbm{1}\big(\textbf{max}\big(\hat{Q}^u_{k,w}) > \tau\big) \mathcal{H}(Q_{k,s}^{u}, \textbf{argmax}(\hat{Q}_{k,w}^u)\big).
\end{gathered}
\end{equation}

Through the aforementioned training process, the general branch and multi-head specialized branch independently handle specific tasks, utilizing a weak-to-strong consistency framework. With each class contributing to the training of other specialized segmentation heads, it fosters a more balanced training process at the pixel level, thereby enhancing the model's segmentation capabilities. \blue{Furthermore, we would like to emphasize that we leverage shared features and introduce only a few MLP-based projectors and segmentation heads, resulting in minimal additional memory overhead during training (only 5.52\% in ACDC) and no extra overhead during inference.}

In summary, we mitigate the scale imbalance in multi-target SSMIS by concurrently training both the general branch and the multi-head specialized branch. In a segmentation task involving $K$ target classes, it is presumed that the individual proportions of these $K$ target classes within the foreground are denoted as $p_i$:
$$p_i = \frac{\sum \mathbbm{1}(Y = i)}{\sum \mathbbm{1}(Y \neq 0)},$$ where $i \in \{1,2,..., K\}$ and $Y$ is the ground truth pixel labels of each image.  
Therefore, $p_i$ should satisfy the following condition: $$\sum_{i=1}^K p_i = 1,$$ \blue{as the sum of the proportions of all categories should be $1$.}

Conventional approaches depict the participation of each class at the pixel level as $p_1, p_2, ..., p_K$. This leads to a scale imbalance caused by differences between $p_i$ and $p_j$, where $i, j \in \{1, 2, ..., K\}$, and $i \neq j$.

\blue{It can be proved that:
\begin{equation}
    |p_i'- \frac{1}{K}| \leq |p_i-\frac{1}{K}|,
\end{equation} where $p'_i$ is the training participation ratio in our method.
This indicates that after our divide-and-conquer strategy, a more balanced training is achieved.}

\blue{In the proposed multi-head specialized branch, the $i$-th target participates as the \textbf{\blue{remaining} classes} in the training of the remaining $K-1$ classes.} More specifically, the participation ratio of the $j$-th class as the \textbf{\blue{remaining} classes} in training is denoted by $1-p_j$. Therefore, it can be deduced that under our framework, the proportion of the $i$-th target participating in the training within the $j$-th specialized segmentation head $P[i,j]$:
$$
P[i,j] = 
\begin{cases}
    p_i & \text{if } i=j \\
    1-p_j & \text{otherwise}
\end{cases}
$$

The total proportion of the $i$-th class $p_i'$ participating in training in our method is obtained as:

$$
p_i'=\frac{\sum_{s=1}^KP[i,s]}{\sum_{s=1}^K\sum_{t=1}^KP[s,t]},
$$\blue{where the numerator is the sum of the proportion in which a single category participates as the target class in its corresponding segmentation head and the proportion in which it participates as the remaining class in other segmentation heads.}

\blue{By substituting the formula for $P[i,j]$ and simplifying, we arrive at the following expression:}
$$
\Rightarrow p_i'- \frac{1}{K}= \frac{2}{(K(K-2)+2)}(p_i-\frac{1}{K}).
$$
\blue{Therefore, 1) when $p_i \neq \frac{1}{K}$:}
$$
\frac{|p_i'- \frac{1}{K}|}{|p_i-\frac{1}{K}|}= \frac{2}{(K(K-2)+2)} < 1.
$$
\blue{It can be obtained by calculation and simplification that:}
$$
|p_i'- \frac{1}{K}| < |p_i-\frac{1}{K}|.
$$
\blue{2) When $p_i = \frac{1}{K}$:}
$$
    p'_i = p_i = \frac{1}{K}.
$$
In summary, $$|p_i'- \frac{1}{K}| \leq |p_i-\frac{1}{K}|,$$ if and only if $p_i=\frac{1}{K},$ the equation holds true.

Note that, $p_i- \frac{1}{K}$ and $p_i'- \frac{1}{K}$ share the same sign (positive/negative). $|p_i- \frac{1}{K}|$ and $|p_i'- \frac{1}{K}|$ represent the distance between the scale proportion of the $i$-th class and the mean value. In this way, \textbf{our design fosters a more balanced training process, resulting in enhanced segmentation performance}.

\subsection{Cross-Branch Consistency}

\begin{table*}[t]
\scriptsize
\centering
\caption{Comparisons with state-of-the-art SSMIS methods on the ACDC dataset.}
%\vspace{-10pt}
\setlength{\tabcolsep}{3mm}{
\resizebox{\linewidth}{!}{
    \begin{tabular}{c|c|cccc|cccc}
    \toprule
    \textbf{Labeled Scans}   & \multirow{2}{*}{\textbf{Venue}} &  \multicolumn{4}{c}{3 Labeled (5\%)}  & \multicolumn{4}{|c}{7 Labeled (10\%)}  \\
    \cmidrule(lr){1-1}
    \cmidrule(lr){3-6}
    \cmidrule(lr){7-10}
    \textbf{Method}       &    &\textbf{DSC} $\uparrow$    &\textbf{Jaccard}  $\uparrow$&\textbf{95HD} $\downarrow$ & \textbf{ASD} $\downarrow$ &\textbf{DSC} $\uparrow$   & \textbf{Jaccard} $\uparrow$ &\textbf{95HD} $\downarrow$  & \textbf{ASD} $\downarrow$\\
    \midrule
    Labeled-only  &  - & 47.83 & 37.01   & 31.16 & 12.62 & 79.41 & 68.11   & 9.35 & 2.70 \\
    \textcolor{gray}{Fully-supervised} & - & \textcolor{gray}{91.44}        & \textcolor{gray}{84.59}   & \textcolor{gray}{1.90} & \textcolor{gray}{0.54}& \textcolor{gray}{91.44}        & \textcolor{gray}{84.59}   & \textcolor{gray}{1.90} & \textcolor{gray}{0.54} \\
    \midrule
    UA-MT~\cite{yu2019uncertainty}    &   MICCAI'19     & 46.04 & 35.97   & 20.08 & 7.75 & 81.65 & 70.64   & 6.88 & 2.02 \\
    FixMatch~\cite{sohn2020fixmatch}    &  NeurIPS’20 &   87.27 &   79.10 &   3.42  &   1.15 &   88.89 &   80.67 &  4.84  & 1.37\\
    SASSNet~\cite{li2020shape}    &   MICCAI'20  &   57.77 &   46.14 &   20.05  &   6.06 &   84.50 &  74.34 &  5.42  &  1.86 \\
    DTC~\cite{luo2021semi}        &  AAAI'21   &   56.90 &  45.67 &   23.36 &   7.39 & 84.29 & 73.92 &  12.81 & 4.01\\
    URPC~\cite{luo2021efficient}     &    MICCAI'21   &  55.87 & 44.64 &  13.60  &  3.74 &   83.10 &   72.41 &   4.84  &  1.53 \\
    MC-Net~\cite{wu2021semi}     &   MICCAI'21  &   62.85 &  52.29 &  7.62  &   2.33  &   86.44 &  77.04 &  5.50  &  1.84\\
    SS-Net~\cite{wu2022exploring}     &   MICCAI'22  &  65.83 &  55.38 &  6.67  &  2.28 &  86.78 & 77.67 &  6.07  & 1.40\\
    CT-CT~\cite{luo2022semi}     &  MIDL'22    &   72.26 &   62.12 &   5.75  &   1.22 &   86.36 & 76.81 &  7.53   &  2.24 \\
    BCP~\cite{bai2023bidirectional}     &  CVPR'23   &   87.59 &   78.67 &   \textbf{1.90}   &   0.67&   88.84 &  80.62 & 3.98   & 1.17 \\
    CPSCauSSL~\cite{miao2023caussl}   &  ICCV'23  &   63.89 &   53.52 &   10.97  &   2.57 & 85.25 & 75.31 &  6.05 & 1.97  \\
    \blue{MSKD~\cite{tu2023semi}}   &  \blue{PRCV'23}  &   \blue{59.41} &   \blue{48.10} &   \blue{39.72}  &   \blue{15.55} & \blue{80.05} & \blue{79.04} &  \blue{14.19} & \blue{2.94}  \\
    \midrule
    CGS (Ours)& This Paper & \BLUE{\textbf{88.83\stdev{0.14}}} &   \BLUE{\textbf{80.62\stdev{1.36}}} &   \BLUE{{2.42\stdev{0.08}}}   &   \BLUE{\textbf{0.66\stdev{0.07}}} & \BLUE{\textbf{89.83\stdev{0.11}}} &   \BLUE{\textbf{82.11\stdev{1.64}}} &   \BLUE{\textbf{2.08\stdev{0.07}}}  &  \BLUE{\textbf{0.68\stdev{0.04}}} \\
    \bottomrule
    \end{tabular}
    }
}
\vspace{-10pt}
\label{tab:acdc}
\end{table*}

To strengthen the interrelationship between the general branch and the multi-head specialized branch, we integrate these branches by cross-branch consistency losses.
On one hand, each branch generates pseudo-labels using unlabeled data for the other branch, facilitating interaction between them. Consequently, we define the bidirectional consistency losses as follows:
\begin{equation}
\label{eq11}
    \mathcal{L}_{c1} = \mathcal{H}(P_s^u, B), \quad \mathcal{L}_{c2} = \frac{1}{K}\sum_{k=1}^K \mathcal{H}(Q^u_{k,s}, B_k),
\end{equation}
% \begin{equation}
%     \mathcal{L}_{G \rightarrow S} = \frac{1}{K}\sum_{k=1}^K \mathcal{H}(Q^u_{k,s}, B_k),
% \end{equation}
where $B$ is the ensemble pseudo-label from all specialized segmentation heads and $B_k$ is the pseudo-label for the $k$-th specialized segmentation head from the general branch.

On the other hand, we calculate a mask, denoted as $M_C$, for the regions where both branches reach consensus. \blue{This directs attention to areas where the the predictions of the both branches align, thereby improving segmentation robustness and accuracy.} In regions where consensus is not reached between the branches, additional training is undertaken in subsequent iterations to prevent error accumulation and potential degradation in the training process. The consensus-based loss optimizes both the general branch and the multi-head specialized branch concurrently. We define the consensus-based consistency loss as follows:
% \begin{equation}
% \begin{split}
%    \mathcal{L}_c =  &\mathcal{H}(P_s^u, B) + \frac{1}{K}\sum_{k=1}^K \mathcal{H}(Q_s^{k,u}, B^k) \\
%  +&M \odot \mathcal{H}(P_s^u, A)) + \frac{1}{K}\sum_{k=1}^K M \odot \mathcal{H}(Q_s^{k,u}, A^k)
% \end{split}
% \end{equation}
\begin{equation}
    \mathcal{L}_{c3} = M_C \odot \mathcal{H}(P_s^u, A) + \frac{1}{K}\sum_{k=1}^K M_C \odot \mathcal{H}(Q_s^{k,u}, A^k),
\end{equation}
where $C$ denote consensus, $\odot$ represents the multiplication operation, $A$ represents the ensemble pseudo-label from the specialized branch, $A^k$ is the pseudo-label for the $k$-th specialized segmentation head. Therefore, the consistency loss can be summarized as:
\begin{equation}
    \mathcal{L}_c = \mathcal{L}_{c1} + \mathcal{L}_{c2} + \mathcal{L}_{c3},
\end{equation}
and the total loss can be summarized as:
\begin{equation}
    \mathcal{L} = \mathcal{L}_{sup} + \mathcal{L}_{sup}^\text{MH} + \lambda( \mathcal{L}_u +  \mathcal{L}_u^\text{MH}) + \mathcal{L}_c,
\end{equation}
where $\lambda$ is the hyper-parameter to balance losses.

The general branch benefits from its capacity to entirely concentrate on inter-class relationships, while the multi-head specialized branch effectively addresses scale imbalance. Therefore, their complementary interaction enhances the overall model performance. 
This straightforward adjustment consistently leads to notable enhancements over the initial weak-to-strong consistency framework. 
% This suggests that our proposed training approach empowers the model with enhanced feature extraction capabilities.

\subsection{Inter-Head Error Detection}
Building upon the multi-head specialized branch mentioned above, we further explore the relationships among multiple segmentation heads. We observe that, even though these segmentation heads focus on distinct tasks, they are required to achieve pixel-level consistency with specific rules. For example, consider two specialized heads, $\textbf{Head}_i$ and $\textbf{Head}_j$. In the segmentation task of $\textbf{Head}_i$, the pixels of \textbf{target class} belong to the pixels of \textbf{\blue{remaining} classes} in the segmentation task of $\textbf{Head}_j$, and vice versa. Therefore, only the following two cases indicate a high likelihood that the pixel is correctly identified:

\begin{itemize}[leftmargin=10pt]
    \item \emph{Case 1: Each of the specialized segmentation heads identifies this pixel as \textbf{background}.}
    \item \emph{Case 2: A specific segmentation head indicates this pixel as the \textbf{target class} exclusively when the remaining heads identify it as the \textbf{\blue{remaining} classes}.}
\end{itemize}
% \begin{algorithm}[t]
% \caption{\small{Pseudocode of IHED in a PyTorch-like style.}}
% \label{alg:sd}
% \definecolor{codepurple}{rgb}{0.4,0.2,0.6}
% \definecolor{codegreen}{rgb}{0.1,0.6,0.1}
% \lstset{
%   backgroundcolor=\color{white},
%   basicstyle=\fontsize{9pt}{9pt}\ttfamily\selectfont,
%   columns=fullflexible,
%   breaklines=true,
%   captionpos=b,
%   commentstyle=\fontsize{9pt}{9pt}\color{codegreen},
%   keywordstyle=\fontsize{9pt}{9pt}\color{blue},
% }
% \begin{lstlisting}[language=python]
% # K: number of specialized segmentation heads
% # ps: pseudo-labels from the K branches 
% # (0: background, 1: target and 2: \blue{remaining})

% matrix = torch.stack(ps, dim=0)
% Detec_Matrix = torch.sum(matrix, dim=0)
% # Case 1: all of them are background
% mask1 = (Detec_Matrix == 0)
% # Case 2: target and (K-1) \blue{remaining}
% mask2 = (Detec_Matrix == (2*K-1))
% return mask1 | mask2
% \end{lstlisting}
% \end{algorithm}

Therefore, we introduce the Inter-Head Error Detection (IHED) module to enhance the accuracy of pseudo-labels produced by the teacher network by filtering out misidentified pixels. Within this module, we examine all the pseudo-labels generated by the specialized heads to validate the accuracy of each pixel assignment. Specifically, for the $k$-th specialized segmentation head, a pixel belongs to class $k$ only if the remaining $K - 1$ branches classify it as part of their respective \textbf{\blue{remaining} classes}. In this meaning, we obtain the error detection matrix $M_d$.
% The corresponding pseudocode is provided in Algorithm~\ref{alg:sd} and gets the error detection matrix $M_d$. 
We can update the $B$ in Eq.~(\ref{eq11}) as follows:

\begin{equation}
    B' = M_d \odot B.
\end{equation}

% In Fig.~\ref{fig:dice_pseudo}, it is evident that our Inter-Head Error Detection method achieves notably superior accuracy in excluding erroneous pixels on the SegTHOR dataset, compared to employing solely a fixed confidence threshold.

\subsection{Inference}
\blue{During the training phase, we integrate $K$ additional specialized segmentation heads on UNet. We note that once the model achieves convergence, the segmentation performance achieved by the mixing outcomes from the $K$ specialized segmentation heads closely aligns with that of the general branch (see Section ~\ref{sec:ablation}). Consequently, specialized segmentation heads can be discarded in inference. Therefore, the proposed method does not incur additional memory and time costs during inference.}

\begin{table*}[t]
\scriptsize
\centering
\caption{Comparisons with state-of-the-art SSMIS methods on the SegTHOR dataset.}
\setlength{\tabcolsep}{3.5mm}{
\resizebox{\linewidth}{!}{
    \begin{tabular}{c|c|cccc|cccc}
    \toprule
    \textbf{Labeled Scans}   & \multirow{2}{*}{\textbf{Venue}} &  \multicolumn{4}{c}{3 Labeled (10\%)}  & \multicolumn{4}{|c}{6 Labeled (20\%)}  \\
    \cmidrule(lr){1-1}
    \cmidrule(lr){3-6}
    \cmidrule(lr){7-10}
    \textbf{Method}       &    &\textbf{DSC} $\uparrow$    &\textbf{Jaccard}  $\uparrow$&\textbf{95HD} $\downarrow$ & \textbf{ASD}$\downarrow$ &\textbf{DSC} $\uparrow$   & \textbf{Jaccard}$\uparrow$ &\textbf{95HD}$\downarrow$  & \textbf{ASD} $\downarrow$\\\midrule
    Labeled-only  & - & 73.10 & 59.70   & 34.23 & 12.39 & 76.98 & 64.64   & 21.16 & 5.76 \\
    \textcolor{gray}{Fully-supervised} & - & \textcolor{gray}{84.67} & \textcolor{gray}{74.73}   & \textcolor{gray}{4.67} & \textcolor{gray}{1.62} & \textcolor{gray}{84.67} & \textcolor{gray}{74.73} & \textcolor{gray}{4.67} & \textcolor{gray}{1.62} \\
    \midrule
    UA-MT~\cite{yu2019uncertainty} & MICCAI'19 & 75.67 & 62.55 & 17.15 & 6.51 & 77.03 & 64.91 & 23.16 & 6.70 \\
    FixMatch~\cite{sohn2020fixmatch} & NeurIPS'20 & 79.38 & 68.52 & 6.95 & 2.25 & 81.94 & 72.40 & 6.84 & 1.91 \\
    URPC~\cite{luo2021efficient} & MICCAI'21 & 75.15 & 62.38 & 29.94 & 8.22 & 77.52 & 65.56 & 18.75 & 6.17 \\
    SS-Net~\cite{wu2022exploring} & MICCAI'22 & 74.85 & 62.56 & 14.25 & 3.58 & 77.98 & 66.43 & 13.89 & 3.32 \\
    CPS~\cite{chen2021semi} & CVPR'21 & 73.51 & 60.41 & 34.22 & 9.69 & 77.62 & 66.63 & 12.64 & 4.65 \\
    CT-CT~\cite{luo2022semi} & MIDL'22 & 75.33 & 62.27 & 22.50 & 6.10 & 78.72 & 66.71 & 15.95 & 4.64 \\
    DHC~\cite{wang2023dhc} & MICCAI'23 & 74.68 & 62.14 & 27.89 & 3.88 & 77.85 & 63.15 & 25.03 & 3.01 \\
    BCP~\cite{bai2023bidirectional} & CVPR'23 & 78.69 & 66.67 & 39.60 & 7.74 & 79.06 & 67.11 & 39.58 & 3.06 \\
    MagicNet~\cite{chen2023magicnet} & CVPR'23 & 73.83 & 61.26 & 28.71 & 3.76 & 76.32 & 64.01 & 24.07 & 2.92 \\
    CPSCauSSL~\cite{miao2023caussl} & ICCV'23 & 75.60 & 63.22 & 19.36 & 5.47 & 78.07 & 66.37 & 14.63 & 5.16 \\
    \blue{MSKD~\cite{tu2023semi}}   &  \blue{PRCV'23}  &   \blue{74.23} &   \blue{62.06} &   \blue{20.69}  &   \blue{4.49} & \blue{78.12} & \blue{64.48} &  \blue{14.23} & \blue{4.98}  \\
    \midrule
    CGS (Ours) & This Paper & \BLUE{\textbf{81.74\stdev{1.29}}} & \BLUE{\textbf{71.66\stdev{3.14}}} & \BLUE{\textbf{6.46\stdev{0.14}}} & \BLUE{\textbf{2.05\stdev{0.09}}} & \BLUE{\textbf{83.69\stdev{1.45}}} & \BLUE{\textbf{74.52\stdev{2.57}}} & \BLUE{\textbf{5.09\stdev{0.11}}} & \BLUE{\textbf{1.79\stdev{0.10}}} \\
    \bottomrule
    \end{tabular}
}
}
\vspace{-10pt}
\label{tab:segthor}
\end{table*}

\section{Experiment}
\blue{In this section, we conduct experiments to compare our method with existing approaches, showcasing its effectiveness. Furthermore, we perform ablation studies and further analysis to assess the contribution of each individual module in our approach.}
\subsection{Dataset}
\BLUE{The experiments on the ACDC and SegTHOR datasets are designed to evaluate performance of our method on simple tasks with a large number of scans and few categories, while the experiment on the Synapse dataset assesses the effectiveness of our method with more categories. }

\textbf{ACDC dataset~\cite{bernard2018deep}.} ACDC dataset is a three-target (\emph{i.e.}, right ventricle, left ventricle and myocardium) segmentation dataset, containing scans of 100 patients. We strictly follow the setting used in \cite{ssl4mis2020}. \blue{For each image, we resize it to $224\times224$ before feeding it into the network. The data set is divided into training, validation, and test sets with a ratio of 7:1:2.}

\textbf{SegTHOR dataset~\cite{lambert2020segthor}.} Segmentation of THoracic Organs (SegTHOR) addresses the problem of Organs at Risk (OAR) segmentation in CT images. The goal of the SegTHOR challenge is to automatically segment four OAR targets: heart, aorta, trachea, and esophagus. 
\blue{
The dataset is partitioned into training, validation, and testing sets, comprising 28, 4, and 8 volumes, respectively. Each 3D volume is converted into 2D slices and resized to dimensions of $224 \times 224$.}

\begin{table*}[t]
\centering
\scriptsize
\caption{Comparison with state-of-the-art SSMIS methods on the Synapse dataset with 3 labeled volumes}
\label{tab:synapse}
\setlength{\tabcolsep}{3mm}{
\resizebox{\linewidth}{!}{
\begin{tabular}{c|c|c|c|cccccccc}
\toprule
\multirow{2}{*}{\textbf{Methods}} & \multirow{2}{*}{\textbf{Venue}} & \multirow{2}{*}{\textbf{Avg. DSC $\uparrow$}} & \multirow{2}{*}{\textbf{Avg. HD95 $\downarrow$}} & \multicolumn{8}{c}{\textbf{DSC $\uparrow
$ of Each Class}} \\
\cmidrule(lr){5-12}
 &  &  &  & \textbf{Ao.} & \textbf{Ga.} & \textbf{LK.} & \textbf{RK.} & \textbf{Li.} & \textbf{Pa.} & \textbf{Sp.} & \textbf{St.}  \\
\midrule
Labeled-only & - &  39.00 & 89.22 & 62.39 & 26.10 & 26.19 & 38.11 & 80.66 & 15.32 & 42.57 & 20.70\\
\textcolor{gray}{Fully-supervised} & - & 
\textcolor{gray}{73.65}  & \textcolor{gray}{57.39} &\textcolor{gray}{87.30} &\textcolor{gray}{59.80} &\textcolor{gray}{75.39} &\textcolor{gray}{67.70} &\textcolor{gray}{91.37} &\textcolor{gray}{52.21} &\textcolor{gray}{83.77} &\textcolor{gray}{71.64} \\
\midrule
UA-MT~\cite{yu2019uncertainty} & MICCAI'19 & 38.02 & 82.82 & 65.52 & 22.98 & 27.66 & 32.04 & 79.14 & 17.87 & 43.28 & 15.68\\
FixMatch~\cite{sohn2020fixmatch} & NeurIPS'20 & 58.33 & 80.79 & 82.16 & 19.45 & 71.71 & 65.49 & 89.80 & 29.23 & \textbf{69.15} & 40.38\\
URPC~\cite{luo2021efficient} & MICCAI'21 & 39.29 & 80.60 & 65.15 & 28.37 & 30.04 & 42.06 & 75.25 & 16.36 & 36.95 & 20.16 \\
SS-Net~\cite{wu2022exploring}& MICCAI'22 & 44.05 & 64.41 & 78.11 & 28.95 & 35.09 & 43.07 & 79.49 & 16.29 & 48.06 & 23.37\\
CPS~\cite{chen2021semi}& CVPR'21 & 39.90 & 72.01 &  61.30 & 29.24 & 24.86 & 39.71 & 80.86 & 18.17 & 45.79 & 19.20\\
CT-CT~\cite{luo2022semi}& MIDL'22 & 55.08 & 81.79 & 75.90 & - & \textbf{68.00} & 66.48 & 89.38 & 38.73 & 68.55 & 32.90 \\
BCP~\cite{bai2023bidirectional}& CVPR'23 & 55.97 & 87.46 & 79.07 & 16.21 & 67.72 & 67.01 & 84.68 & \textbf{30.15} & 61.48 & 41.46 \\
CPSCauSSL~\cite{miao2023caussl}& ICCV'23 & 41.55 & 84.20 & 64.86 & \textbf{35.42} & 28.32 & 43.90 & 80.60 & 19.31 & 38.72 & 21.24\\
\midrule
CGS (Ours) & This Paper & \BLUE{\textbf{60.20\stdev{0.37}}} & \BLUE{\textbf{61.40\stdev{1.04}}} & \textbf{82.36} & 32.00 & 65.39 & \textbf{72.33} & \textbf{91.35} & 29.49 & 61.89 & \textbf{46.77}\\
\bottomrule
\end{tabular}
}
}
% \vspace{-10pt}
\end{table*}

\textbf{Synapse dataset.} Synapse\footnote{\url{https://www.synapse.org/Synapse:syn3193805/wiki/217789}} dataset consists of 30 computed tomography (CT) scans annotated with
eight abdominal organs, where class imbalance occurs frequently. Following \cite{zhang2023customized}, we adopt 18 and 12 cases for training and testing, respectively. \blue{Similar to ACDC and SegTHOR, each slice is resized to $224\times224$.}

\subsection{Setting}
\textbf{Evaluation Metrics. }We adopt four evaluation metrics: 3D Dice Similarity Coefficient (DSC \%), Jaccard Score (\%), 95\% Hausdorff Distance (95HD) in the voxel, and Average Surface Distance (ASD) in the voxel. Dice and Jaccard metrics primarily assess the degree of overlap between two object regions. 
ASD is an indicator for calculating the average distance between boundaries, while 95HD quantifies the closest point distance between them.

\textbf{Training.} All experimental procedures are conducted using PyTorch on a single Nvidia RTX 2080Ti GPU with 11 GB of memory. For training the segmentation models, we employ a Stochastic Gradient Descent (SGD) optimizer with a momentum of 0.9 and weight decay of 0.0001. The backbone consists of UNet with several specialized heads, equivalent to UNet during inference. The batch size is set to 24, comprising 12 labeled slices and 12 unlabeled slices. The setting for the hyperparameter $\lambda$ follows a ramp-up policy~\cite{laine2016temporal} with a maximum value of 2.0. For all the datasets, the threshold $\tau$ is set to 0.9. The training iteration is set to 30000, and the initial learning rate is established at 0.01 for all experiments. \blue{Additionally, different data augmentation methods are employed to enable a more equitable comparison with previous methods. We apply cropping, rotation and flip operations to generate weakly augmented data, while color jitter and CutMix are employed to obtain strongly augmented images} \blue{to enhance the robustness of the model}. The probability of CutMix is set to $1.0$ for ACDC and SegTHOR, while $0.5$ for Synapse. In addition, take ACDC as an example, our method comprises 1.8160M parameters, slightly surpassing UNet's parameter count of 1.8138M (increased by 0.12\%).

\textbf{Testing.} 
For the ACDC and SegTHOR datasets, we select the model that achieves the highest performance on the validation set as the final model. 
For the Synapse dataset, the model saved at the final iteration is chosen as the final model. 
\BLUE{We conduct experiments using three different seeds on three datasets and report their mean values and standard deviations, demonstrating that our method consistently achieves stable performance improvements.}

\blue{\textbf{Number of branches.} We list the number of branches (see Tab.~\ref{number_of_b}) of our method on the three datasets in the main experiments. During training, we add an additional lightweight segmentation head, while we use only a single segmentation head during inference.}

 \begin{figure*}[h]
    \centering
    %\vspace{-20pt}
    \includegraphics[width=0.95\linewidth]{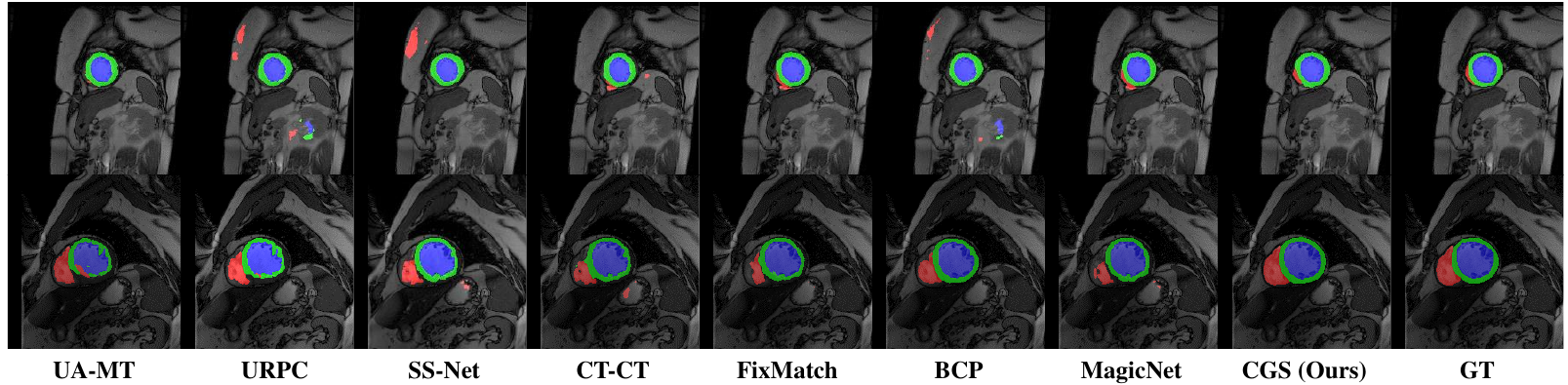}
    % \captionsetup{justification=centering}
    \vspace{-5pt}
    \centering
    \caption{
Visualization of segmentation results on ACDC. Our method achieves segmentation results that most closely match the ground truth.}
    \vspace{-10pt}
    \label{fig:vis}
\end{figure*}

\begin{table}[h]
\vspace{-10pt}
\centering
\setlength\tabcolsep{6.5mm}
{
\caption{The number of branches used during training and inference.}
\resizebox{\linewidth}{!}{
\begin{tabular}{c|ccc}
\toprule
Stage   & ACDC & SegTHOR & Synapse\\
\midrule
Training   & 4 & 5 &  9\\
Inference    & 1 & 1 & 1\\
\bottomrule
\end{tabular}%
\label{number_of_b}
}
}
\vspace{-20pt}
\end{table}

\begin{figure}[b]
    \centering
    \vspace{-10pt}
    \includegraphics[width=0.8\linewidth]{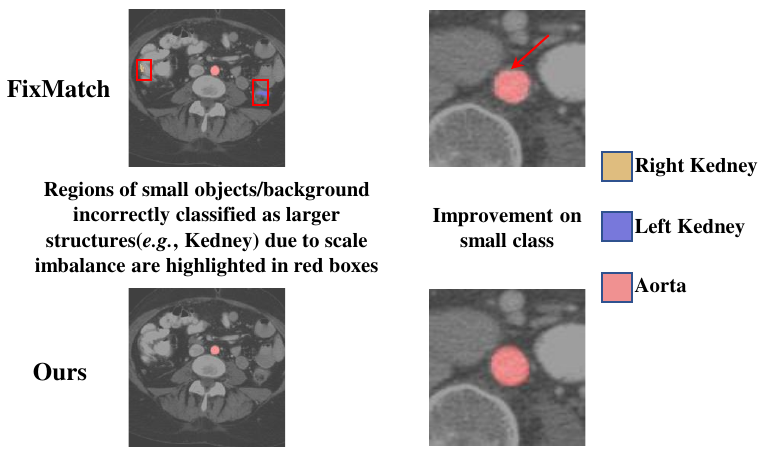}
    \vspace{-10pt}
    \caption{\BLUE{Segmentation results on Synapse. Our method achieves more precise and distinct boundary segmentation for small objects and reduces misclassification as larger structures (\emph{e.g.}, Kedney) caused by scale imbalance.}} 
    \label{fig:synapse}
    % \vspace{-10pt}
\end{figure}

\subsection{Compared with SOTA Methods}
We compare our method with various baselines: UA-MT \cite{yu2019uncertainty}, FixMatch~\cite{sohn2020fixmatch}, SASSNet \cite{li2020shape}, DTC \cite{luo2021semi}, URPC \cite{luo2021efficient}, MC-Net \cite{wu2021semi}, SS-Net \cite{wu2022exploring}, CT-CT \cite{luo2022semi}, CPS \cite{chen2021semi},  BCP \cite{bai2023bidirectional}, CPSCauSSL~\cite{miao2023caussl}, DHC~\cite{wang2023dhc} and MagicNet \cite{chen2023magicnet}. DHC is specifically designed for handling class imbalance, and MagicNet is tailored for multi-organ segmentation. Some methods (\emph{i.e.}, MagicNet, DHC, \emph{etc}) are devised specifically for 3D datasets. To ensure a fair comparison with our approach, we have crafted a 3D dataset derived from the SegTHOR dataset. \blue{Moreover, we compare our methods with \cite{tu2023semi}, which focus on multi-scale at the feature level
of the image.} \blue{Nearly all semi-supervised methods for 2D medical image segmentation, including our approach, utilize the UNet architecture as the backbone. In addition, we introduce $K$ small branches, one for each category, where $K$ represents the number of categories.}

Furthermore, we observe that FixMatch, a classic semi-supervised learning technique, continues to exhibit exceptional performance in medical image segmentation tasks. Therefore, FixMatch stands out as an outstanding baseline for further research in SSMIS. Moreover, the performance of semi-supervised segmentation closely rivals that of fully supervised learning when a large number of labeled samples are available.

\begin{figure*}[h]
    \centering
    \begin{minipage}[b]{0.32\textwidth}
        \centering
        \includegraphics[width=0.8\linewidth]{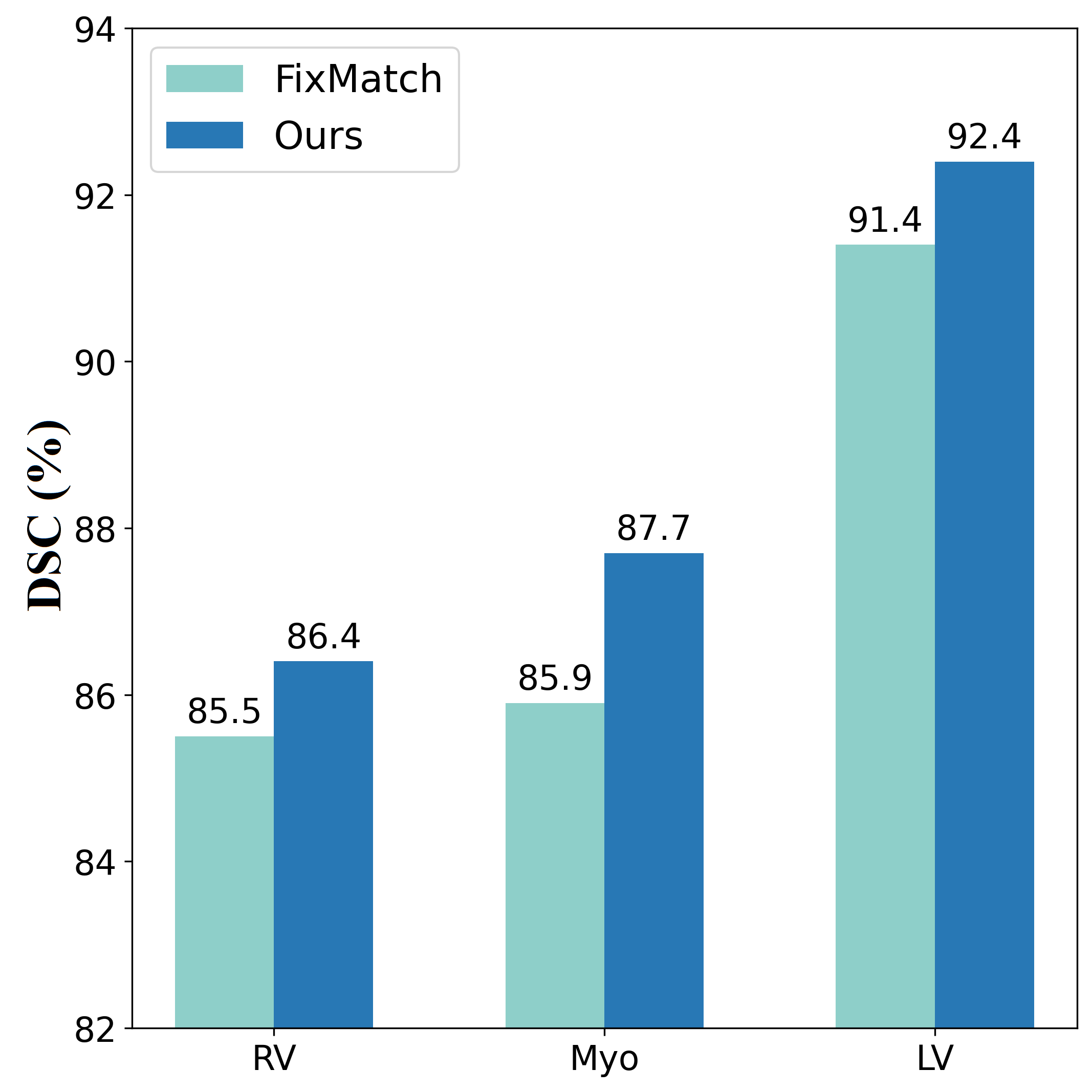}
        \caption{\blue{Comparison (DSC \%) with FixMatch on ACDC for each class.}}
        \label{fig6}
    \end{minipage}
    \hfill
    \begin{minipage}[b]{0.32\textwidth}
        \centering
        \includegraphics[width=0.8\linewidth]{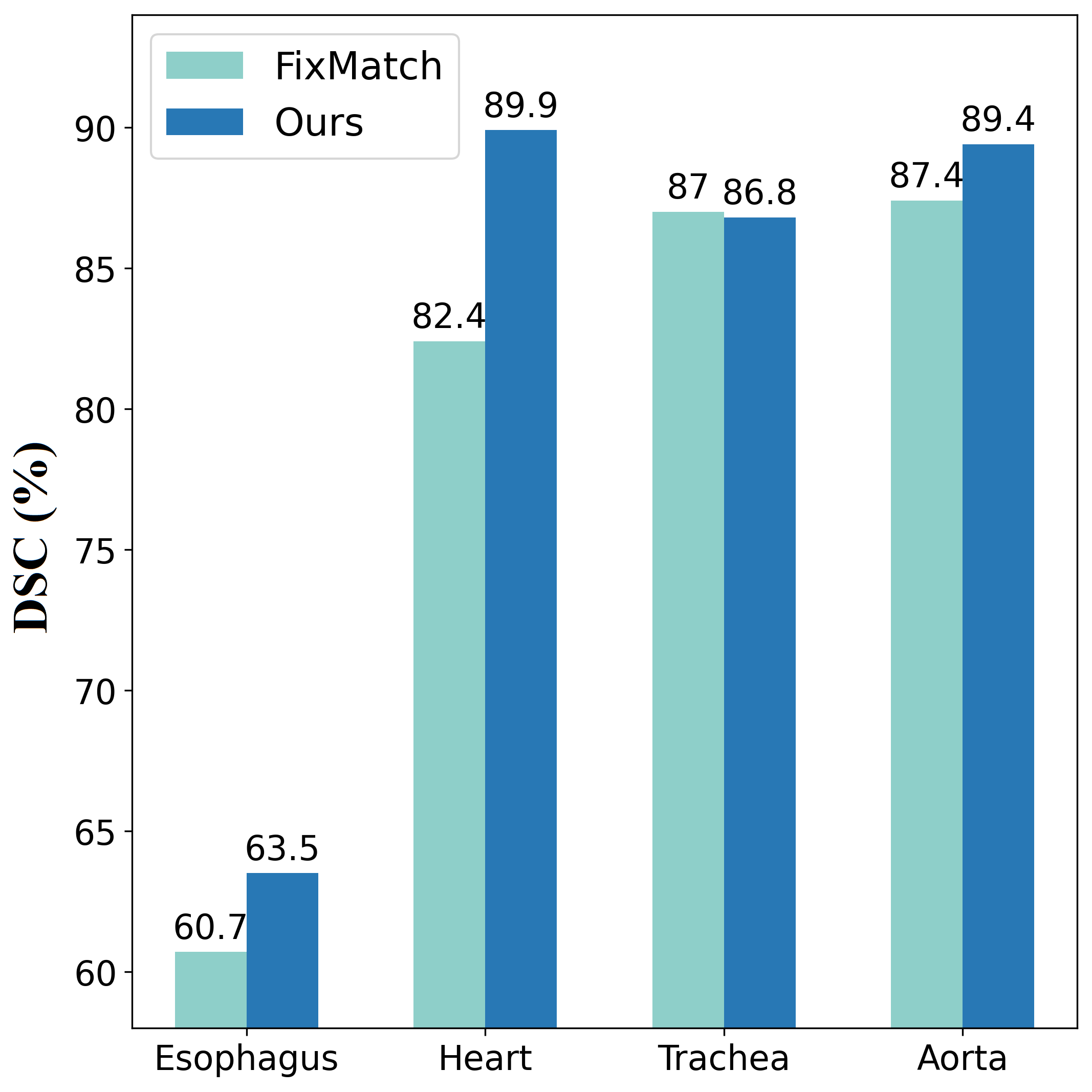}
        \caption{\blue{Comparison (DSC \%) with FixMatch on SegTHOR for each class.}}
        \label{fig7}
    \end{minipage}
    \hfill
    \begin{minipage}[b]{0.32\textwidth}
        \centering
        \includegraphics[width=0.8\linewidth]{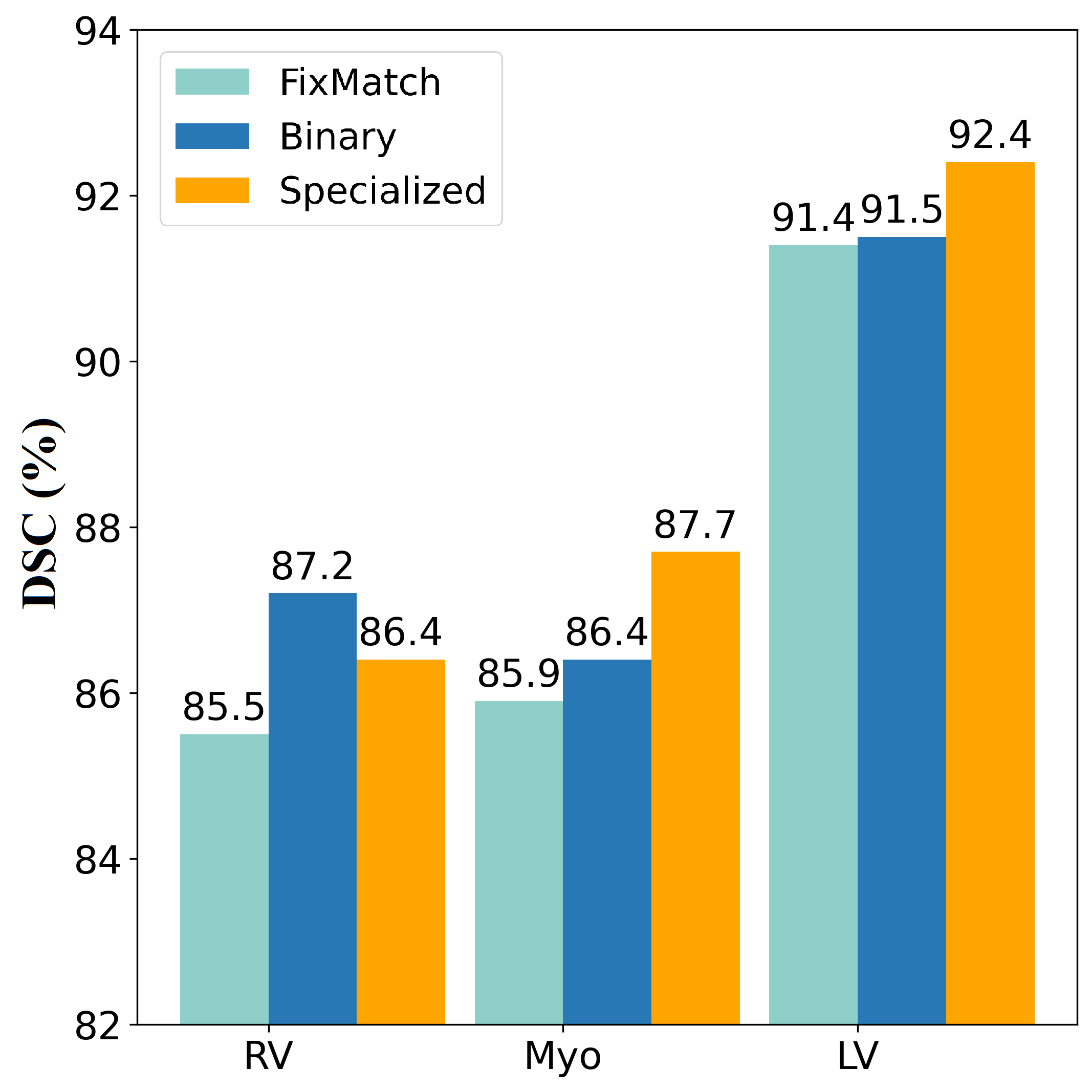}
        \label{fig8}
        \caption{Performance (DSC \%) of FixMatch, binary segmentation heads, and the proposed specialized segmentation heads on each class.}
    \end{minipage}
\vspace{-10pt}
\end{figure*}

\textbf{Results on ACDC Dataset.} Tab.~\ref{tab:acdc} illustrates the averaged performance of four-class segmentation outcomes on ACDC, utilizing labeled ratios of 5\% and 10\%. Our proposed CGS surpasses all state-of-the-art methods in terms of Dice, Jaccard, and ASD while being marginally less effective than the latest BCP regarding 95HD. This suggests that our method outperforms the state-of-the-art approaches in pixel-level predictions and achieves a comparable boundary accuracy to BCP. \BLUE{Segmentation examples on ACDC are depicted in Fig.~\ref{fig:vis}. The segmentation results produced by our method align most closely with the ground truth.}

\textbf{Results on SegTHOR Dataset.} 
On the SegTHOR dataset, we present experimental results with labeled data proportions of 10\% and 20\%. As indicated in Tab. \ref{tab:segthor}, our approach excels across all evaluation metrics, surpassing other methods by a significant margin.

\textbf{Results on Synapse Dataset.} 
The Synapse dataset exhibits pronounced class imbalance. To assess the efficacy of our approach on datasets with such imbalances, we conducted comparisons with SOTA methods on this dataset. In Tab.~\ref{tab:synapse}, we showcase the experimental outcomes derived from the Synapse dataset, employing three labeled volumes. Our proposed method surpasses other SOTA methods in both DSC and HD95, notably evident in challenging targets such as the right kidney (RK). \blue{The segmentation performance of small organ (Ao.) on Synapse is shown in Fig.~\ref{fig:synapse}, where we can see that our proposed method offers a significant advantage in the segmentation of small objects and
mitigates the occurrence of misclassification as large objects, compared with baseline FixMatch.
}
% \begin{wrapfigure}{r}{1.5in}
%     \centering
%     \vspace{-10pt}
%     \includegraphics[width=0.82\linewidth]{figs/synapse.png}
%     % \vspace{-10pt}
%     \caption{\blue{Performance of small organ on Synapse.}}
%     \label{fig:synapse}
%     \vspace{-10pt}
% \end{wrapfigure}
This outcome strongly supports our insight of addressing scale imbalance among different organs. Furthermore, we compared our method with two SAM-based segmentation models, AutoSAM~\cite{hu2023efficiently} and SAM~\cite{chen2023sam} Adapter on Synapse.
Although AutoSAM and SAM Adaptor perform well on general datasets, its results fall short of ours in specific medical tasks. For example, they only achieved the DSC (\%) of 55.69 and 28.28, respectively.
Additionally, it is worth noting that our method has fewer parameters than theirs.

\begin{table}[h]
% \vspace{-10pt}
\caption{Ablation study (DSC \%) on ACDC and SegTHOR.}
\vspace{-10pt}
\renewcommand\arraystretch{1}
\setlength\tabcolsep{4.5pt} % Adjust the value as needed
\centering
{\small
\resizebox{\linewidth}{!}{
\begin{tabular}{c|cccc|cc}
\noalign{\smallskip}
\hline
\noalign{\smallskip}
Methods & BSH & SSH  & $\mathcal{L}_c$ & IHED  & ACDC (5\%) & SegTHOR (10\%)   \\ 
\noalign{\smallskip}
\hline
\noalign{\smallskip}
\#1 & & & & &  87.27 & 79.38  \\ 
\#2 &\checkmark & & & &  87.74 & 79.56 \\ 
\#3 &\checkmark & &\checkmark & &  87.89 & 80.24 \\ 
\#4 & &\checkmark & & & 87.79 & 80.32  \\ 
\#5 & &\checkmark &\checkmark & &  88.36  & 81.19  \\ 
\#6 & &\checkmark & &\checkmark &  88.17 & 80.65 \\ 
\noalign{\smallskip}
\hline
\noalign{\smallskip}
\#7 & &\checkmark&\checkmark &\checkmark & \textbf{88.83} &  \textbf{81.74} \\
\noalign{\smallskip}
\hline
\noalign{\smallskip}
\end{tabular}}}
% %\vspace{-5pt}
\label{tab:ablation}
% \vspace{-10pt}
\end{table}

\subsection{Ablation Study and Further Analysis}
\label{sec:ablation}

\textbf{Effectiveness of each component.} 
We conducted ablation studies to evaluate the individual contributions of two key components: Inter-Head Error Detection (IHED) and Cross-Branch Consistency ($\mathcal{L}_c$). These components are tested within both the Binary Segmentation Head (BSH), which segments only the background and target class, and the Specialized Segmentation Head (SSH), which segments extra \blue{remaining} classes. \blue{Both Inter-Head Error Detection (IHED) and Cross-Branch Consistency ($\mathcal{L}_c$) are built upon the Specialized Segmentation Head (SSH). As a result, it is not feasible to conduct independent ablation experiments for IHED and $\mathcal{L}_c$. The evaluation was performed across the ACDC and SegTHOR datasets with 3 labeled scans.} The results, detailed in Tab.~\ref{tab:ablation}, indicate that both BSH and our proposed SSH exhibit enhanced segmentation performance for the main branch, even in the absence of consistency loss from the two-branch framework. In addition, the performance of our proposed SSH surpasses that of the standard BSH in identical setups. Furthermore, the inclusion of other components has similarly contributed to the overall improvement of our method. 

\textbf{Quantitative analysis for the confidence threshold $\tau$.} 
\begin{wrapfigure}{r}{1.6in}
    \centering
    % \vspace{-10pt}
    \includegraphics[width=0.82\linewidth]{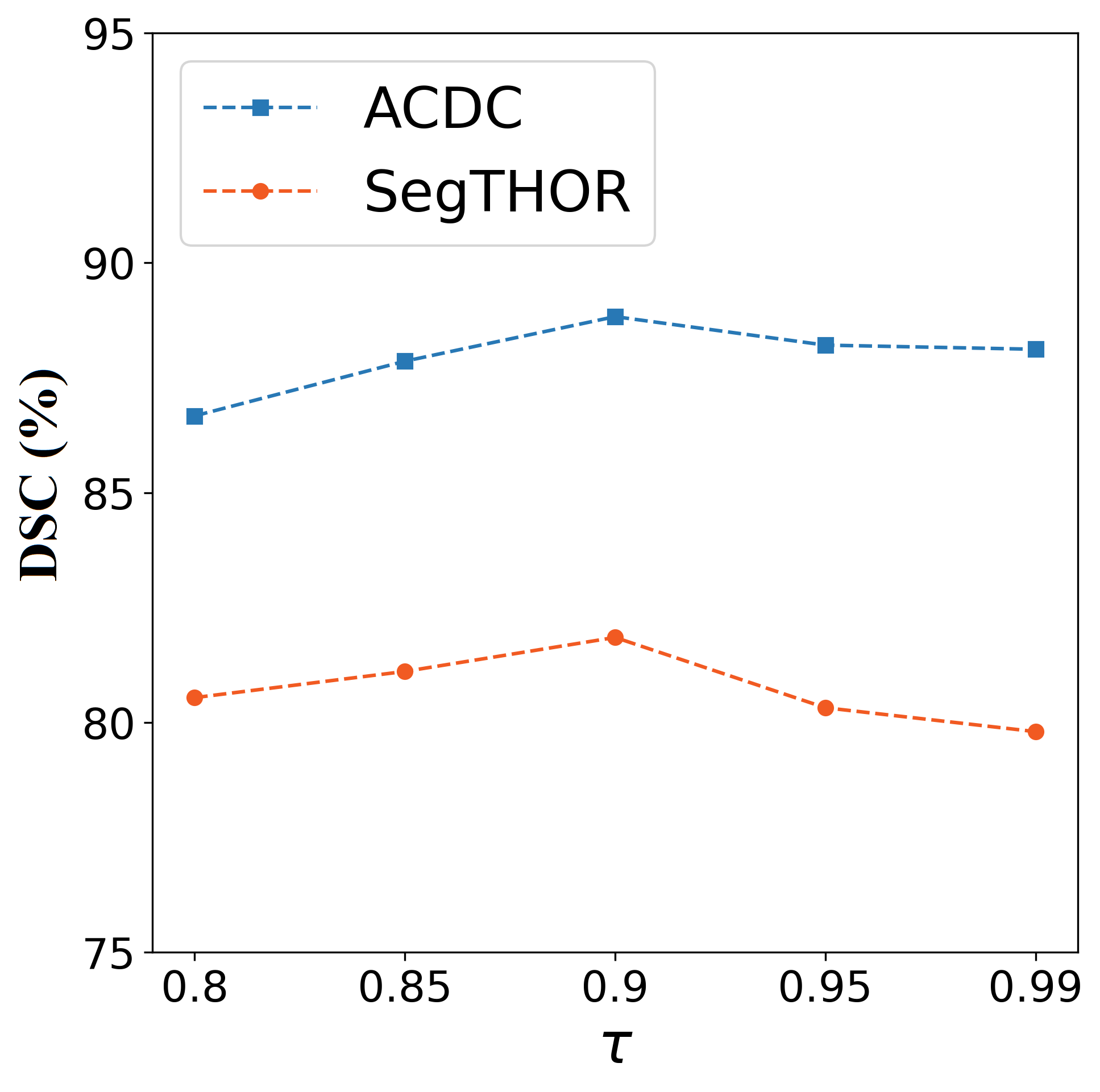}
    \vspace{-10pt}
    \caption{Performance (DSC) with different $\tau$ on ACDC and SegTHOR.}
    \label{tau}
    \vspace{-10pt}
\end{wrapfigure}
We conduct ablation experiment of $\tau$ on the ACDC (5\% labeled) and SegTHOR (10\% labeled) datasets, as depicted in Fig.~\ref{tau}. The optimal value of $\tau$ identified for both datasets is 0.9. This indicates that higher values of $\tau$ diminish the available data, whereas lower values tend to increase the generation of incorrect pseudo-labels.

\textbf{Performances without CutMix.} 
To demonstrate that the effectiveness of our approach extends beyond the use of strong augmentations like CutMix, we intentionally exclude CutMix from our experiments, and compare our proposed method against the strong baseline FixMatch \cite{sohn2020fixmatch}. The results, detailed in Tab.~\ref{ablation6}, reveal that even without CutMix, our proposed method achieves a notably greater improvement. This reinforces the efficacy and robustness of our approach.

\begin{table}[h]
% \vspace{-10pt}
\caption{Comparison (DSC \%) with FixMatch without CutMix.}
\centering
\scriptsize
\setlength\tabcolsep{3.0mm}
\resizebox{\linewidth}{!}{%
\begin{tabular}{c|c|c|cc}
\toprule
Method & Dataset & Labeled ratio  & \multicolumn{2}{c}{DSC (\%)} \\
\midrule
FixMatch \cite{sohn2020fixmatch} & \multirow{2}{*}{ACDC} &  \multirow{2}{*}{5\%}  & 84.19 \\
Ours &  &   & 87.38 \\
\midrule
FixMatch \cite{sohn2020fixmatch} & \multirow{2}{*}{SegTHOR} &  \multirow{2}{*}{10\%}  & 76.92 \\
Ours &  &   & 79.33 \\
\bottomrule
\end{tabular}
}
\label{ablation6}
% \vspace{-10pt}
\end{table}

\textbf{Performance of our method on \blue{other baselines}. }
\blue{To showcase the versatility of our proposed approach, we employ the CGS training strategy within the CPS~\cite{chen2021semi} and MT~\cite{tarvainen2017mean} on both the ACDC (5\% labeled) and SegTHOR (10\% labeled) datasets.} As shown in Tab.~\ref{cps}, our method achieved substantial improvements across different datasets. \blue{It is worth mentioning that our approach can be applied as a plug-in to most existing state-of-the-art semi-supervised medical image segmentation methods.}

\begin{table}[h]
\centering
\scriptsize
\setlength\tabcolsep{3.2mm}
\vspace{-10pt}
\caption{\blue{Performance (DSC \%) as a plugin on CPS and MT.}}
\resizebox{\linewidth}{!}{%
\begin{tabular}{c|c|c|cc}
\toprule
Method & Dataset & Labeled ratio  & \multicolumn{2}{c}{DSC (\%)} \\
\midrule
CPS & \multirow{4}{*}{ACDC} &  \multirow{4}{*}{5\%}  & 73.51 \\
CPS + CGS &  &   & 75.83 \\
\blue{MT} &  &   & \blue{50.30} \\
\blue{MT + CGS} &  &   & \blue{53.21} \\
\midrule
CPS & \multirow{4}{*}{SegTHOR} &  \multirow{4}{*}{10\%}  & 77.62 \\
CPS + CGS &  &   & 79.15 \\
\blue{MT}  &  &   & \blue{68.48} \\
\blue{MT + CGS} &  &   & \blue{70.73} \\
\bottomrule
\end{tabular}
}
\label{cps}
% \vspace{-10pt}
\end{table}

\textbf{Depth of projector. }  
In our method, we incorporate a projector into each specialized segmentation head to enhance the diversity of different branches. Nonetheless, employing a projector with excessive depth might escalate memory usage and potentially trigger gradient vanishing problems. To identify the optimal depth, we conducted experiments, the results of which are presented in Tab.~\ref{ablation7}. The model attains peak performance when the depth is set to 1.

\begin{table}[h]
\vspace{-5pt}
\caption{Comparison (DSC \%) with different depths of the projector.}
% \vspace{-5pt}
\centering
\scriptsize
\setlength\tabcolsep{2.0mm}
\resizebox{\linewidth}{!}{%
\begin{tabular}{c|c|cccc}
\toprule
Dataset & Labeled ratio & 0 & 1 & 2 & 3 \\
\midrule
ACDC & 5\%  & 87.06 &88.83 &86.72& 87.70\\
SegTHOR & 10\% & 80.38& 81.74 & 79.35& 81.17\\
\bottomrule
\end{tabular}
}
\label{ablation7}
\vspace{-7pt}
\end{table}

\textbf{Performance improvement at each class.} 
In Fig. 8, we present a comparative analysis of the performance between traditional binary segmentation heads and our proposed specialized segmentation heads across different classes. We aim to demonstrate how our CGS leverages the advantages of both the general branch and the multi-head specialized branch. To further elucidate this point, we showcase the experimental results of our method for each class in Fig.~\ref{fig6} and Fig.~\ref{fig7} on ACDC and SegTHOR datasets, each with 3 labeled volumes. The outcomes demonstrate that our CGS strategy improves segmentation performance across the majority of categories.

\blue{
\textbf{Comparision with the straightforward ensemble of generalist heads.}
On the ACDC dataset, we compared the DSC (\%) on different numbers of labeled volumes with those from an ensemble of generalists trained using 4 different seeds. We employ three different ensemble strategies to ensemble these four models. 1) The first approach involves \textbf{averaging (Avg.)} the prediction results from all models 2) The second strategy focuses on selecting the regions with the highest \textbf{confidence (Conf.)} from each model. 3) The third method is a \textbf{voting (Vot.)} mechanism. As shown in the Tab.~\ref{tab:ensemble}, our method consistently outperforms the ensemble approach, demonstrating its superior effectiveness. Ensembling multiple generalists would incur exponential computational overhead, yet our method still achieves superior performance.}

\begin{table}[h]
\centering
% \tiny
\vspace{-10pt}
\setlength\tabcolsep{2.5mm}
{
\caption{\blue{Comparison (DSC \%) with ensemble 4 baseline models.}}
\label{tab:ensemble}
\resizebox{\linewidth}{!}{
\begin{tabular}{c|c|ccc|c}
\toprule
Labeled Volumes  & Mean  & Avg. & Conf. & Vot. & Ours\\
\midrule
3 &  \BLUE{86.86\stdev{1.37}} &  88.35 &  88.14 &  88.28 & \textbf{88.83}\\
7  & \BLUE{88.76\stdev{0.20}} & 89.59 & 89.41 & 89.61 & \textbf{89.83}\\
\bottomrule
\end{tabular}
}%
}
\vspace{-10pt}
\end{table}

\blue{\textbf{Efficiency analysis.} Tab.~\ref{tab9} compares the FLOPs (Floating point operations) and number of parameters of our method with different approaches. While our approach adds a small, manageable increase in FLOPS during training, it maintains the same inference cost.} \BLUE{In addition, Tab.~\ref{tab:tab10} presents an analysis of the memory requirements for our method as the number of classes increases. In this experiment, we set the batch size to 24 and utilized teacher and student networks. The results indicate that while the memory consumption of our method increases slightly with more classes, this overhead is acceptable considering the performance gains achieved by our proposed method.}

\begin{table}[h]
\centering
{
\vspace{-10pt}
\caption{GPU memory (GB) of different number of classes.}
\label{tab:tab10}
\resizebox{\linewidth}{!}{
\begin{tabular}{c|cccccc}
\toprule
Number of classes   & 3  & 4 & 5 & 6 & 7 & 8\\
\midrule
GPU memory (GB)   & 7.13 & 7.78 & 8.58 & 9.36 & 9.85 & 10.62\\
\bottomrule
\end{tabular}%
}
}
\vspace{-10pt}
\end{table}

\BLUE{\textbf{Scalability with higher-resolution image. }To accommodate the higher resolutions, we adjusted the batch size from 24 to 8 and 6. As shown in Tab. \ref{tab:tab11}, our method consistently outperformed the baseline, demonstrating superior performance even with increased resolution. } 

\begin{table}[h]
\centering
\vspace{-10pt}
{
\caption{Comparison (DSC \%) with baseline with increased resolution. }
\label{tab:tab11}
\setlength\tabcolsep{6.5mm}
\resizebox{\linewidth}{!}{
\begin{tabular}{c|c|cc}
\toprule
Dataset&Input Size   & FixMatch  & Ours \\
\midrule
\multirow{3}{*}{ACDC} & 224 & 87.27 & 88.83 \\
& 400 & 84.85& 86.35 \\
& 512 & 84.37& 85.70 \\
\bottomrule
\end{tabular}%
}
}
\vspace{-20pt}
\end{table}

\begin{table}[h]
\centering
\caption{\blue{Efficiency analysis of the
different approaches on GFLOPs.}}
\setlength\tabcolsep{2.6mm}
\resizebox{\linewidth}{!}{
{
\begin{tabular}{c|ccc|c|c}
\toprule
\blue{Stage} & \blue{UA-MT} & \blue{BCP}  & \blue{URPC} & \blue{FixMatch} & \blue{Ours} \\
\midrule
\blue{Training} & \blue{2.99} & \blue{2.99}  & \blue{3.02} & \blue{2.99} & \blue{3.14}\\
\blue{Inference} & \blue{2.99} & \blue{2.99}  & \blue{2.99} & \blue{2.99} & \blue{2.99}\\
\bottomrule
\end{tabular}%
}
}
\label{tab9}
\vspace{-10pt}
\end{table}

\blue{\textbf{Experiments conducted using TransUNet~\cite{chen2021transunet} as the backbone.} All the compared SOTA methods, including ours, use the UNet~\cite{ronneberger2015u} as the backbone for 2D medical image segmentation. In addition, we have added an experimental comparison of our method and FixMatch with the backbone replaced by TransUNet~\cite{chen2021transunet} in Tab.~\ref{transunet}. 
Even with TransUNet as the backbone, our method continues to show significant improvement over the baseline, demonstrating its robustness and effectiveness across different architectures.}

\begin{table}[h]
\centering
{
\vspace{-10pt}
\caption{The DSC (\%) on ACDC and SegTHOR using TransUNet as the backbone.}
\setlength\tabcolsep{5.1mm}
\resizebox{\linewidth}{!}{%
\begin{tabular}{c|c|cc}
\toprule
Dataset&Labeled Volumes   & FixMatch  & Ours \\
\midrule
\multirow{2}{*}{ACDC} & 3 & 76.80 & 81.92 \\
& 7 & 80.62& 83.72 \\
\midrule
\multirow{2}{*}{SegTHOR} & 3  & 63.34 & 65.68\\
& 7 & 72.47 & 74.80\\
\bottomrule
\end{tabular}%
}
\label{transunet}
}
\vspace{-10pt}
\end{table}

\textbf{The feasibility of discarding multi-head specialized branch.} During the inference phase, we perform experiments by integrating segmentation results from both the multi-head specialized branch and the general branch at different ratios. We observe that this does not affect the inference outcomes (refer to Tab.~\ref{ablation5}). Consequently, it is feasible to completely exclude the multi-head specialized branch during inference, leading to a reduction in memory usage.

\begin{table}[h]
\centering
\setlength\tabcolsep{2.0mm}
\vspace{-10pt}
\caption{Comparison (DSC \%) with different mixing ratios.}
\resizebox{\linewidth}{!}{%
\begin{tabular}{c|c|ccccc}
\toprule
Dataset & Labeled ratio  & 0 & 0.2 & 0.5 & 0.8 &1.0\\
\midrule
ACDC &5\%  & 88.83 & 88.81 & 88.81 & 88.82 & 88.83\\
ACDC &10\%  & 89.82&89.83&89.82&89.82&89.83\\\midrule
SegTHOR &10\% & 81.74&81.73&81.73&81.74&81.72 \\
SegTHOR &20\% & 83.69&83.66&83.65&83.68&83.69 \\
\bottomrule
\end{tabular}%
}
% \vspace{-14pt}
\label{ablation5}
\end{table}

\blue{
\textbf{Experiments handling occurrence imbalance.} For the Synapse dataset, although both issues are present, we aim to demonstrate that scale imbalance has a more
pronounced impact on performance. We consider the Gallbladder (Ga.) and Pancreas (Pa.), which are target classes with significantly lower occurrence frequencies compared to other classes. As shown in Tab.~\ref{resampling}, existing methods (\emph{e.g.}, FixMatch~\cite{sohn2020fixmatch}) achieve significantly poorer results on these two less frequent classes. We attempt to mitigate the impact of occurrence imbalance by applying resampling on these two classes. Although the resampling strategy aimed at alleviating occurrence imbalance can improve performance on Ga. and Pa., it may lead to performance degradation on other classes. Consequently, its overall average performance across all classes remains inferior to our method CGS, which focuses on addressing scale imbalance. These experimental results clearly indicate that scale imbalance is a more significant issue than occurrence imbalance in Synapse. Moreover, the experimental results demonstrate that, by addressing scale imbalance as a foundation, CGS can be combined with resampling to mitigate occurrence imbalance, thereby further improving the performance. In future research, we are committed to developing a more comprehensive approach to better address both occurrence and scale imbalance simultaneously.}

\begin{table}[h]
\centering
\scriptsize
\vspace{-10pt}
\caption{\blue{Performance (DSC \%) on Synapse. RS. denotes resampling.}}
\resizebox{\linewidth}{!}{%
\label{resampling}
\setlength{\tabcolsep}{1.5mm}{
{
\begin{tabular}{c|c|cc}
\toprule
\multirow{2}{*}{Methods}  & \multirow{2}{*}{Avg. DSC $\uparrow$}  & \multicolumn{2}{c}{DSC $\uparrow
$ of Each Class} \\
\cmidrule(lr){3-4}
 &     & \textbf{Ga.} & \textbf{Pa.}  \\
\midrule
FixMatch~\cite{sohn2020fixmatch}   & 58.33   & 19.45 & 29.23 \\
FixMatch~\cite{sohn2020fixmatch} + RS.   & 59.06  & 39.61 & 37.00 \\
\midrule
CGS (Ours)   & 60.20 & 32.00 & 29.49 \\
CGS (Ours) + RS.    & 64.93  & 53.25   & 35.80 \\
\bottomrule
\end{tabular}
}
}
}
\end{table}

\section{Conclusion}
In this study, we investigate the challenge of scale imbalance in multi-target SSMIS. To tackle this issue, we present the CGS framework, which includes a general branch and a multi-head specialized branch. Building upon this dual-branch framework, we incorporate cross-branch consistency losses for mutual learning and implement the inter-head error detection module to enhance the quality of pseudo-labels. 
Our method demonstrates excellent performance with a few number of segmentation categories, as evidenced by results from ACDC, SegTHOR, and Synapse datasets. The proposed method performs well on datasets with few categories (ACDC/SegTHOR) as well as on datasets with more categories (Synapse).
Furthermore, our method can integrate as a modular component within existing SSL methods, \emph{e.g.}, CPS.

\section{Acknowledgments}
\BLUE{This work was supported by NSFC Project (62222604, 62206052, 624B2063), China Postdoctoral Science Foundation (2024M750424), Fundamental Research Funds for the Central Universities (020214380120, 020214380128), State Key Laboratory Fund (ZZKT2024A14), Postdoctoral Fellowship Program of CPSF (GZC20240252), Jiangsu Funding Program for Excellent Postdoctoral Talent (2024ZB242), Jiangsu Science and Technology Major Project (BG2024031), and Shandong Natural Science Foundation (ZR2023MF037).}

\bibliographystyle{IEEEtran}
\bibliography{ref}

\end{document}